\documentclass{article}

\usepackage{microtype}
\usepackage{graphicx}
\usepackage{booktabs} %
\usepackage{hyperref}

\usepackage[accepted]{icml2026}

\usepackage{amsmath}
\usepackage{amssymb}
\usepackage{mathtools}
\usepackage{amsthm}

\usepackage{amsmath,amsfonts,bm}

\def\eqref#1{equation~\ref{#1}}

\def\1{\bm{1}}

\def\vtheta{{\bm{\theta}}}

\def\vx{{\bm{x}}}

\def\vz{{\bm{z}}}

\def\mA{{\bm{A}}}

\def\mG{{\bm{G}}}

\DeclareMathAlphabet{\mathsfit}{\encodingdefault}{\sfdefault}{m}{sl}
\SetMathAlphabet{\mathsfit}{bold}{\encodingdefault}{\sfdefault}{bx}{n}

\def\gD{{\mathcal{D}}}

\def\gL{{\mathcal{L}}}

\newcommand{\vt}{\bm{\tau}}
\newcommand{\vth}{\bm{\theta}}

\usepackage{cleveref}
\usepackage{subcaption}
\usepackage{amsmath, amssymb, bm}
\usepackage{mathtools}
\usepackage{amsthm}
\usepackage{longtable}
\usepackage{makecell}

\usepackage[table]{xcolor}
\usepackage{tcolorbox}

\newcommand{\todores}[1]{\textcolor{red}{$00.00$}}

\definecolor{lightorange}{HTML}{FFBE88}
\definecolor{lightblue}{HTML}{94DDFD}

\usepackage{booktabs}

\usepackage{nicefrac}
\usepackage{enumitem}

\usepackage{wrapfig}

\newcommand{\norm}[1]{\left\lVert #1 \right\rVert_2^2}

\usepackage{algorithm}       %

\usepackage{multirow}
\usepackage{bbm}

\definecolor{forestGreen}{RGB}{34, 139, 34}
\definecolor{firebrick}{RGB}{178, 34, 34}

\newcommand{\jac}{\mathrm{J}}

\newcommand{\nicepar}[1]{\noindent \textbf{#1}}

\usepackage{xspace}
\newcommand*{\ie}{\textit{i.e.,\@\xspace}}

\newcommand*{\eg}{\textit{e.g.,\@\xspace}}

\usepackage{pifont}

\newcommand{\cmark}{\textcolor{green!60!black}{\ding{51}}}
\newcommand{\xmark}{\textcolor{red!70!black}{\ding{55}}}

\newlist{tightitemize}{itemize}{3}
\setlist[tightitemize]{
    label=\textbullet, 
    nosep,          
    list
    leftmargin=*,    
    after=\vspace{5pt}
}

\newcommand{\methname}{\texttt{DELTA}}
\newcommand{\methnamecomplete}{DistillEd Linearized Task Arithmetic}

\usepackage{array}
\newcommand{\PreserveBackslash}[1]{\let\temp=\\#1\let\\=\temp}
\newcolumntype{C}[1]{>{\PreserveBackslash\centering}p{#1}}
\newcolumntype{R}[1]{>{\PreserveBackslash\raggedleft}p{#1}}
\newcolumntype{L}[1]{>{\PreserveBackslash\raggedright}p{#1}}

\newif\ifrevision
\revisionfalse   

\ifrevision
  \newcommand{\changed}[1]{\textcolor{blue}{#1}}
\else
  \newcommand{\changed}[1]{#1}
\fi

\Crefname{equation}{Eq.}{Eqs.}
\Crefname{figure}{Fig.}{Figs.}
\Crefname{tabular}{Tab.}{Tabs.}
\Crefname{section}{Sec.}{Secs.}
\Crefname{table}{Tab.}{Tabs.}

\crefname{equation}{Eq.}{Eqs.}
\crefname{figure}{Fig.}{Figs.}
\crefname{tabular}{Tab.}{Tabs.}
\crefname{section}{Sec.}{Secs.}
\crefname{table}{Tab.}{Tabs.}
\crefname{algorithm}{Alg.}{Algs.}

\definecolor{myred}{RGB}{220, 80, 80}       
\definecolor{myredlight}{RGB}{245, 210, 210} 

\definecolor{myblue}{RGB}{70, 130, 180}      
\definecolor{mybluelight}{RGB}{210, 225, 240} 

\newif\iftottfont
\tottfonttrue      %

\newcommand{\tott}[1]{%
  \iftottfont
    \texttt{#1}%
  \else
    #1%
  \fi
}

\theoremstyle{plain}
\newtheorem{theorem}{Theorem}[section]

\theoremstyle{definition}
\newtheorem{definition}[theorem]{Definition}

\theoremstyle{remark}

\icmltitlerunning{Distilling Linearized Behavior for Effective Task Arithmetic}

\begin{document}

\pdfstringdefDisableCommands{%
  \def\\{ }
}

\twocolumn[
  \icmltitle{Distilling Linearized Behavior into Non-Linear Fine-Tuning \\ for Effective Task Arithmetic}

  \icmlsetsymbol{equal}{*}

  \begin{icmlauthorlist}
    \icmlauthor{Thomas Sommariva}{yyy}
    \icmlauthor{Francesca Morandi}{yyy,xxx}
    \icmlauthor{Simone Calderara}{yyy}
    \icmlauthor{Angelo Porrello}{yyy}
  \end{icmlauthorlist}

  \icmlaffiliation{yyy}{AImageLab, Department of Engineering “Enzo Ferrari”, University of Modena and Reggio Emilia, Italy}
  \icmlaffiliation{xxx}{University of Pisa, Italy}

  \icmlcorrespondingauthor{Angelo Porrello}{angelo.porrello@unimore.it}

  \icmlkeywords{task arithmetic, model editing, model merging, model reuse, task vector}

  \vskip 0.3in
]

\printAffiliationsAndNotice{}  %

\begin{abstract}
Task vector composition has emerged as a promising paradigm for editing pre-trained models, enabling model merging through addition and unlearning through subtraction. Fine-tuning in the tangent space of a pre-trained model (\textit{linear fine-tuning}) has proven effective, as it produces task vectors that are naturally disentangled and resistant to interference. However, linearized models suffer from limited expressivity during training and incur higher computational costs at inference time, which restrict their practical applicability.
In this work, we bridge the gap between linear and standard non-linear fine-tuning. We show that linearity with respect to weight perturbations, a property defined in parameter space, can be enforced through constraints in activation space during training. Concretely, we distill hidden representations from a curvature-regularized linearized teacher into a non-linear student trained via conventional fine-tuning. 
We find that the resulting model inherits key properties of linearized models for task arithmetic, enabling effective composition of task vectors and achieving strong performance across vision and language benchmarks without incurring any inference-time overhead.
\end{abstract}

\section{Introduction}
\label{sec:intro}
As deep models continue to grow in scale and complexity, retraining from scratch or even fine-tuning them is becoming increasingly impractical. This trend has motivated a growing body of research toward mechanisms that enable the composition~\cite{liu2023tangent}, the modification~\cite{ilharco2022patching,fierro2025steering}, and reuse~\cite{rinaldi2025update,rinaldi2025gradient,rinaldi2026transporting} of existing models, rather than training new ones. In this context, practitioners can reuse learned capabilities, rapidly customize models, and deploy tailored systems under strict computational or data constraints.

Within this field, task arithmetic~\cite{ilharco2022editing} enables model editing through simple algebraic operations in weight space. Given task-specific models $\{\vth_t = \vth_0 + \vt_t\}_t$ fine-tuned from a common pre-trained model $\vth_0$, the corresponding update vectors $\vt_t$ (\textbf{task vectors}) can be composed to create a single multi-task model (\textit{addition}), or to selectively remove task-specific behaviors (\textit{subtraction}).
\begin{figure}[tb]
  \centering
  \includegraphics[width=0.87\linewidth]{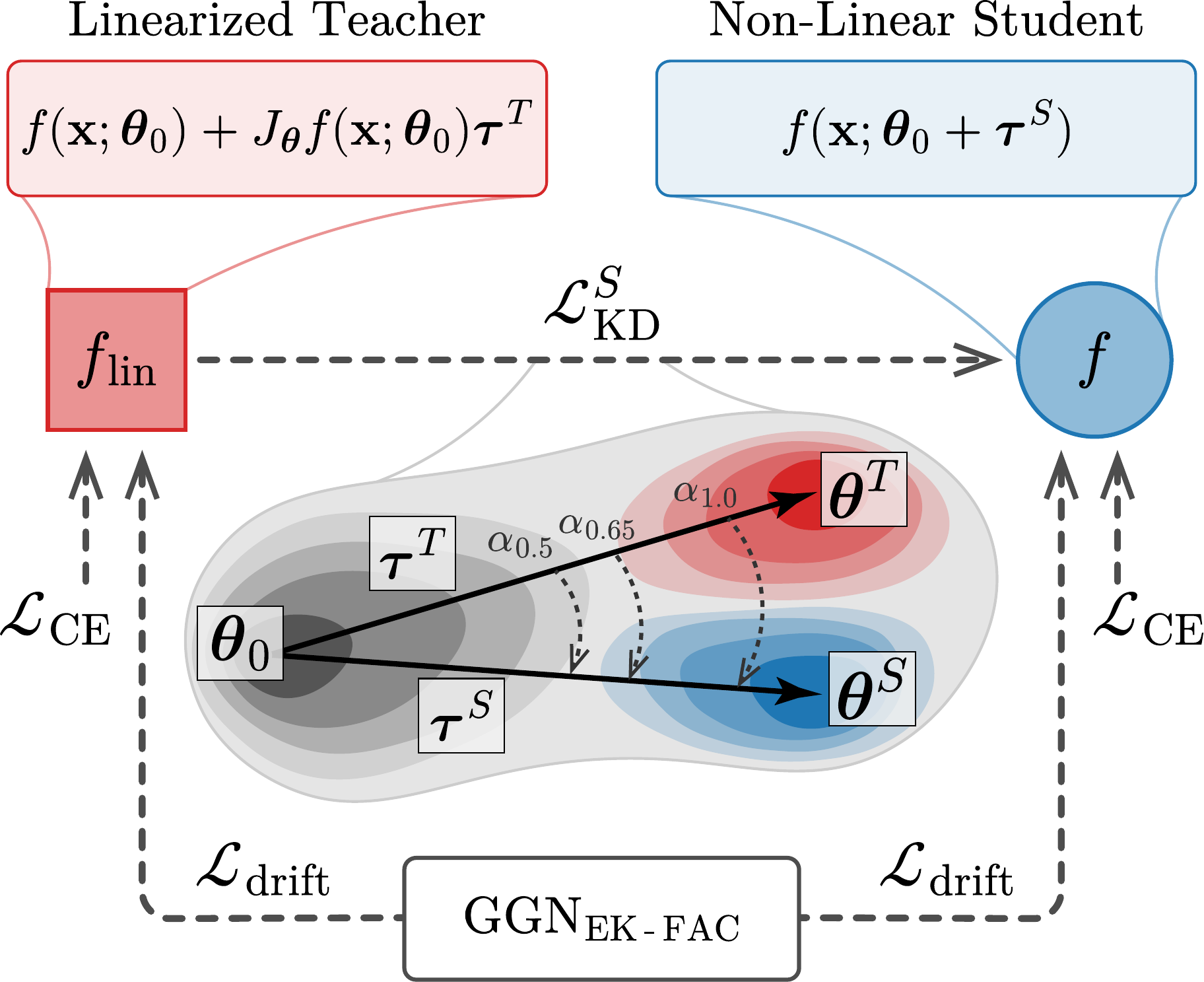}
\caption{\textbf{Overview.} To improve weight disentanglement, a non-linear student model is fine-tuned by distilling a linearized teacher. Both models are trained with curvature-aware regularization based on an approximation of the Generalized Gauss-Newton matrix.}
\vspace{-0.8em}
\label{fig:intro}
\end{figure}

However, the effectiveness of task arithmetic crucially depends on how task vectors are learned. Recent work has shown that fine-tuning in the tangent space of a pre-trained model, commonly referred to as \emph{\textbf{linear fine-tuning}}~\cite{ortiz2023task}, yields task vectors that are more naturally disentangled and substantially less prone to interference -- a property commonly referred to as \textbf{\textit{weight disentanglement}}. Moreover, since models in this regime are linear in weight space, edits in parameter space induce predictable changes in the output space. This property enables the use of explicit and efficient regularization penalties~\cite{yoshidamastering,porrello2025dataless} to further promote disentanglement and task compatibility. In contrast, standard non-linear fine-tuning does not exhibit these properties; as a result, advanced and tailored post-hoc model merging strategies~\cite{gargiulo2025task,marczak2025no,panariello2025accurate,buzzega2025rethinking} are required to mitigate interference effects, with mixed results.

While linearization offers several advantages, there is no free lunch. The computational cost of a single forward pass increases significantly, making deployment more expensive~\cite{ortiz2023task}. Moreover, constraining optimization to the tangent space of the pre-trained model may limit the expressivity of the model. Given the complementary strengths and weaknesses of linearized and standard non-linear training regimes, a natural question arises: 

\begin{tcolorbox}[colback=gray!2!white, colframe=black]
Is there a sweet spot in fine-tuning that mitigates the drawbacks of both linearized and non-linear regimes while preserving \textit{weight disentanglement}, \textit{expressiveness}, and \textit{inference efficiency}?\end{tcolorbox}

We show that linearity with respect to weight perturbations -- a property defined in \textit{parameter space} -- can be induced in a conventional non-linear model by imposing tailored learning objectives in \textit{activation space}. Specifically, we show that linearized behavior and weight disentanglement can be distilled~\cite{hinton2015distilling} by matching the activations of a linearized model. As illustrated in \cref{fig:intro}, we distill intermediate activations from a teacher model trained in tangent space, which guides a student model trained in the standard, non-linear fine-tuning regime. This yields task vectors that can be efficiently composed via addition and subtraction within a standard, deployment-friendly non-linear model.

Building on this insight, we propose \textbf{\methnamecomplete{}} (\textbf{\methname{}}). First, we incorporate curvature-aware regularization~\cite{porrello2025dataless,porrello2025second} to promote disentanglement. Unlike prior work relying on data or statistics from other tasks, we estimate these regularization terms using a third-party \textbf{reference dataset}, yielding a task-agnostic training scheme. Second, instead of distilling from a single teacher--student pair, we sample their weights along the linear path connecting the pre-trained weights to their current values during optimization. This \textbf{along-path distillation} exposes the student to an ensemble of linearly interpolated teachers, enabling a richer approximation of the linearized dynamics and promoting the transfer of linear behavior.

We empirically show that task arithmetic does not require strict linearization, but rather localized and approximately linear update directions. These properties yield a model that preserves the composability of linear fine-tuning while benefiting from the expressivity of standard training; as a result, the student \textbf{outperforms its teacher}. Finally, \methname{} achieves strong performance across vision and language benchmarks and can be applied to generative LLM settings.

\section{Background}
\label{sec:background}
\nicepar{Notation.} Let $f:\mathcal{X}\times\Theta\to\mathbb{R}^d$ be a neural network with $L$ layers, mapping inputs $\vx\in\mathcal{X}\subseteq\mathbb{R}^D$ and weights $\vtheta\in\Theta\subseteq\mathbb{R}^P$ to an intermediate representation in $\mathbb{R}^d$. The final task output is obtained via an additional linear transformation $\phi:\mathbb{R}^d\to\mathcal{Y}\subseteq\mathbb{R}^C$, with the overall model given by $\phi\!\left(f(\vx;\vth)\right)$. We consider a collection of $T$ downstream tasks, where each task $t$ is defined by a triplet $(\mathcal{D}_t,\mu_t,f_t)$: a data support $\mathcal{D}_t\subseteq\mathcal{X}$, an input distribution $\mu_t$ with $\mathrm{supp}(\mu_t)=\mathcal{D}_t$, and a target function $f_t:\mathcal{D}_t\to\mathcal{Y}$.

\nicepar{Task vectors.} For each task $t$, the model is fine-tuned on $\mathcal{D}_t$ starting from the pre-trained weights $\vth_0$, yielding task-specific parameters $\vtheta_t$. The update of the pre-trained model to task $t$ can be represented through the corresponding \emph{task vector} $\vt_t := \vtheta_t - \vth_0$. \emph{Task arithmetic} (TA) \citep{ilharco2022editing} posits that these vectors can be combined via simple linear operations in parameter space to edit model functionality. \emph{Task addition} constructs a multi-task model by linearly combining task vectors, resulting in parameters $\vth_0 + \sum_{t=1}^T \alpha_t \vt_t$. \emph{Task negation} aims to forget a task by subtracting the corresponding task vector from $\vtheta_0$.

A property key to task arithmetic is \emph{\textbf{weight disentanglement}}.
\begin{definition}[\emph{Weight disentanglement -- informal}]
A set of task vectors $\{\vt_t\}_{t=1}^T$ is disentangled if, for each task $t$, applying $\vt_t$ induces negligible changes in the predictions for inputs outside the support of task $t$. In this case, the function $f$ can be decomposed into a sum of spatially localized components vanishing outside a given region.
\end{definition}
\nicepar{Linear fine-tuning.} \citet{ortiz2023task} empirically demonstrated that \textbf{linearized neural networks} exhibit stronger disentanglement than standard non-linear fine-tuning, with improved task arithmetic performance. Formally, a linearized model $f_{\mathrm{lin}}(\vx; \vth)$ is defined via a first-order Taylor expansion around pre-trained parameters $\vtheta_0$:
\begin{equation}
  \label{eq:linearization}
  f_{\mathrm{lin}}(\vx; \vth)
  =
  f(\vx; \vth_0) + \jac_{\vth} f(\vx; \vth_0) (\vth - \vth_0),
\end{equation}
where $\jac_{\vth} f(\vx; \vth_0) \in \mathbb{R}^{d \times P}$ denotes the Jacobian of the model prediction at input $\vx$ evaluated at $\vtheta_0$. 

\nicepar{Regularization in linearized models.} Prior work~\cite{yoshidamastering,porrello2025dataless} showed that even linear fine-tuning admits residual interference between task vectors. A key advantage of the linear regime, however, is that such interference can be analyzed in \emph{closed form}, a property unavailable under standard fine-tuning. In fact, given an example $\vx$ from the dataset $\mathcal{D}_t$, we can compute the \textbf{representation drift}, \ie{} the change of the last layer activation when editing the model with the task vector $\vt_{t'}$:
\begin{equation}
\Delta_{t \to t,t'}(\vx) := \norm{\vz_{t,t'}
- \vz_t}
\propto \smash{\norm{\jac_{\vtheta} f(\vx; \vtheta_0)\, \vt_{t'}}},
\end{equation}
where $\vz_t=f_{\mathrm{lin}}(\vx;\vtheta_0+\alpha\vt_t)$ and $\vz_{t,t'}=f_{\mathrm{lin}}(\vx;\vtheta_0+\alpha\vt_t+\alpha\vt_{t'})$ denote the last-layer representations of $\vx$ before and after the addition of $\vt_{t'}$, respectively. This analytical characterization of interference is exploited by \citet{yoshidamastering} to introduce a regularizer that explicitly penalizes representation drift in linearized models. While effective, this approach requires direct access to the training data of external tasks, which is often restricted by privacy or storage constraints in decentralized settings.
 
In this respect, \citet{porrello2025dataless} showed that the dependence on external task data can be avoided by leveraging the \textbf{generalized Gauss–Newton (GGN) matrix}~\citep{schraudolph2002fast,martens2014new}, a tool widely used in the curvature-aware optimization literature. Under this lens, for linearized models, the representation drift on examples from an external dataset $\gD_t$ has a closed-form quadratic expression:
\begin{equation}
\label{eq:gnn}
    \gL_{\operatorname{drift}}(\vtheta_{t'}) \propto (\vtheta_{t'} - \vtheta_0)^\top \mG_t(\vtheta_0) (\vtheta_{t'} - \vtheta_0) .
\end{equation}
Here, $\mG_t(\vtheta_0) = \frac{1}{|\gD_t|}\sum_{\vx \in \gD_t} \jac_{\vth} f(\vx; \vth_0)^\top \jac_{\vth} f(\vx; \vth_0)$ is the GGN matrix computed on the dataset $\gD_t$, which, once pre-computed, does not require further access to task data. Since the full GGN matrix is intractable -- it scales quadratically with the number of parameters -- \citet{porrello2025dataless} resort to \textit{Kronecker-Factored Approximate Curvature} (KFAC) \citep{martens2015optimizing} -- for a tutorial covering both theory and implementation, see \citet{dangel2025kfac}. KFAC approximates the GGN with a block-diagonal structure, where each layer $l$ is represented as a Kronecker product $\mA^l \otimes \mG^l$, with $\mA^l$ and $\mG^l$ denoting the Gram matrices of the input activations and output gradients, respectively. Notably, incorporating the KFAC approximation into \cref{eq:gnn} enables dataless optimization, as it requires models to share only the Kronecker factors ${\mA^l, \mG^l}$ rather than raw data.
\subsection{Discussion and Limitations}
\label{sec:discussion}
\begin{table}
\centering
\setlength{\tabcolsep}{3pt}
\caption{Comparison of training regimes for task arithmetic.}
\label{tab:regime_comparison}
\resizebox{\linewidth}{!}{%
\begin{tabular}{lccc}
\toprule
 & \textbf{\shortstack{Non-Linear\\FT}} & \textbf{\shortstack{Linear\\FT}} & \textbf{\shortstack{Distilled\\(ours)}} \\
\midrule
Task arithmetic                 & \xmark & \cmark & \cmark \\
Curvature-aware reg.            & \xmark & \cmark & \cmark \\
Robustness to scaling $\alpha$  & \xmark & \cmark & \cmark \\
Expressivity                    & \cmark & \xmark & \cmark \\
Efficiency (inference)          & \cmark & \xmark & \cmark \\
Efficiency (training)           & \cmark & \xmark & \xmark \\
\bottomrule
\end{tabular}
}
\end{table}

To sum up, linearized models offer several \textbf{advantages} for task arithmetic. In particular, they naturally yield task vectors with improved \emph{weight disentanglement}~\cite{ortiz2023task} and admit an exact, closed-form characterization of task interference~\cite{yoshidamastering}. This latter property enables the design of dataless regularizers that minimize \emph{representation drift} through curvature-aware approximations. Finally, linearized models have also been observed~\cite{porrello2025dataless} to be more robust to the choice of scaling coefficients $\{\alpha_t\}_{t=1}^T$, a property that may facilitate deployment without requiring extensive tuning of scaling parameters on a held-out validation set.

Despite these advantages, the linear regime also presents notable \textbf{limitations}. Its reliance on Jacobian-vector products (\cref{eq:linearization}) incurs substantial overhead, doubling the cost during both training and inference (see \cref{app:computational_costs}). Second, constraining the model to remain on the tangent plane around the pre-trained parameters limits expressivity, potentially leading to inferior performance on individual tasks.

Taken together, the drawbacks of the linear regime summarized in \cref{tab:regime_comparison} could hinder its practical deployment in many settings. Hence, our work positions itself in this direction: retaining the inference-time efficiency and flexibility of non-linear fine-tuning, while inducing learning directions compatible with task arithmetic and model merging.

\section{Proposed Method: \methname{}}
\label{sec:method}
\begin{figure*}[tb]
    \centering
    \begin{minipage}[b]{0.495\textwidth}
        \centering
        \includegraphics[width=\linewidth,clip,trim=0 0 0 0]{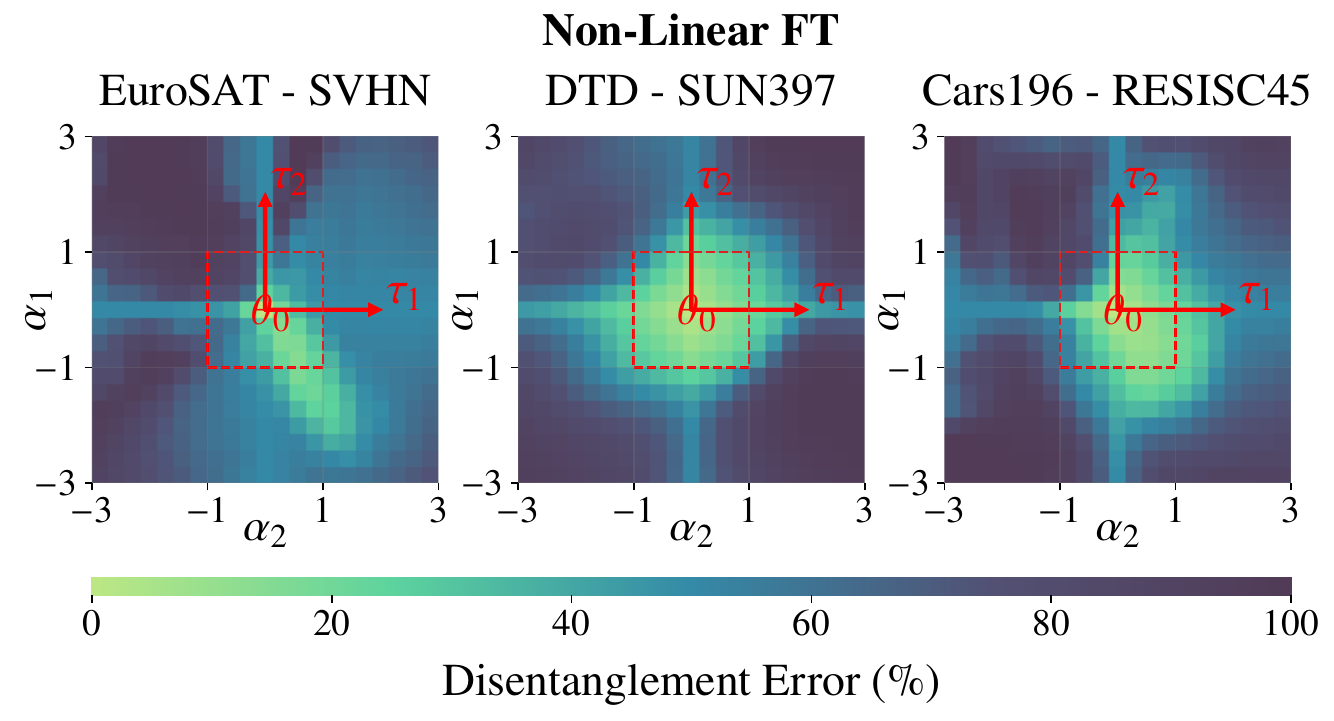}
    \end{minipage}%
    \begin{minipage}[b]{0.495\textwidth}
        \centering
        \includegraphics[width=\linewidth,clip,trim=0 0 0 0]{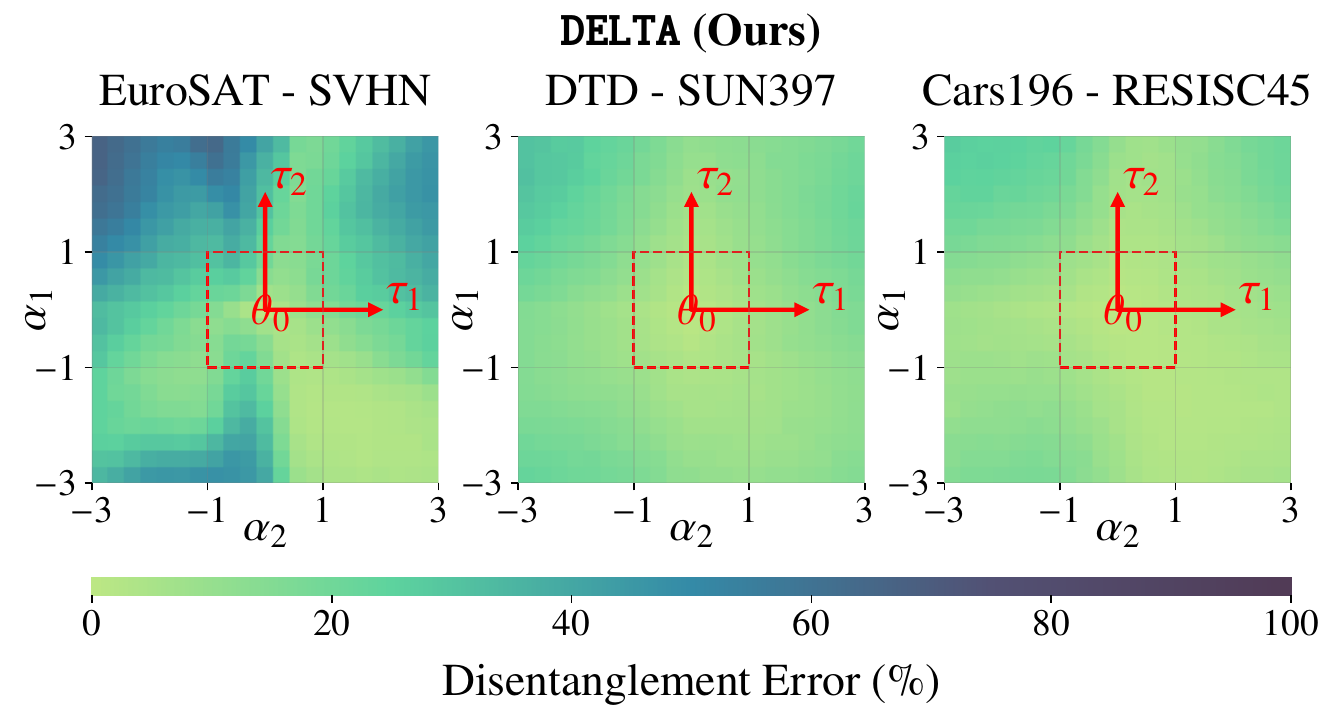}
    \end{minipage}
\caption{The heatmaps show the disentanglement error~\cite{ortiz2023task} of a non-linear CLIP ViT-B/32 (left) and the non-linear distilled student (right) on several task pairs. The light regions denote areas of the weight space where weight disentanglement is stronger.}
\label{fig:disentanglement}
\end{figure*}

Building on the discussion in \cref{sec:discussion}, we seek a training strategy that operates in the standard non-linear setting while retaining the favorable properties of linearized models. To this end, we rely on knowledge distillation~\cite{hinton2015distilling} and propose \textbf{\methnamecomplete{}} (\textbf{\methname{}})\footnote{ \scriptsize{\url{https://github.com/apanariello4/merge-and-rebase}.}}: we train a model via conventional fine-tuning while encouraging it to match the activations of a curvature-regularized \textit{linearized model}, which serves as its teacher.

The key hypothesis is that mimicking the activations of such a teacher biases optimization toward solutions in parameter space that exhibit similar behavior in the student, including linearity to weight perturbations and enhanced weight disentanglement. This is corroborated by \cref{fig:disentanglement}: compared with conventional non-linear fine-tuning, ours greatly reduces the \textbf{disentanglement error}~\cite{ortiz2023task}
\begin{align}
\textstyle
\xi(\alpha_1, \alpha_2) = \sum_{t=1}^2 &\mathbb{E}_{\vx \sim \mu_t} \Big[
\operatorname{dist}\big(
\phi(f(\vx;\vth_0+\alpha_t\vt_t)), \notag \\
&\phi(f(\vx;\vth_0+\alpha_1\vt_1+\alpha_2\vt_2))
\big)
\Big],
\end{align}
\ie{} the discrepancy between the predictions of merged and individual models, with
$\operatorname{dist}(y_1,y_2)=\mathbbm{1}\{y_1 \neq y_2\}$.
\subsection{Teacher–Student Training Setup}
\label{sec:setup}
For each task $t=1,\dots,T$, we consider two models built upon the \textbf{same pre-trained initialization} $\vth_0$: a non-linear student $f(\vx;\vth_t^S)$ and a linearized teacher $f_{\operatorname{lin}}(\vx;\vth_t^T)$. The teacher corresponds to the first-order linearization of the model around $\vth_0$ (see \cref{eq:linearization}), while the student operates in the standard non-linear regime. We define the corresponding task vectors as $\vt_t^S = \vth_t^S - \vth_0$ and $\vt_t^T = \vth_t^T - \vth_0$.

\nicepar{Online feature-level distillation.} The two models are trained jointly in an \textbf{online fashion}, eliminating the need for separate training stages and yielding a single, unified optimization process. Furthermore, distillation is enforced directly in the \textbf{feature space}: concretely, we align the activations of the last hidden layer before the final projection head, encouraging the non-linear student to match the linear representations produced by the teacher. As a distillation criterion, we adopt a mean squared error (MSE) loss between teacher and student features, which will later be generalized to a novel proposed along-path distillation objective.

In the following sections, we detail the loss functions applied to the teacher (\cref{sec:teacher}) and the student (\cref{sec:student}).
\subsection{Training the Linearized Teacher}
\label{sec:teacher}
As in standard distillation frameworks, the linearized teacher is first trained to solve the task under consideration. Accordingly, its most immediate learning signal is the task loss itself (\eg{} cross-entropy for classification). However, to act as an effective teacher in the context of task arithmetic, we promote \textbf{weight disentanglement} through curvature-aware regularization. As discussed in \cref{sec:background}, operating in the linear regime enables explicit, closed-form control of representation drift, thereby encouraging task-specific update directions that are well separated and suited for task arithmetic. We can thus summarize the twofold objective of the teacher as a composite loss \(\gL_t^T(\mathcal{X};\vth_t^T)\), defined as:
\begin{tcolorbox}[colback=blue!5!gray!10,boxsep=1pt, left=1pt, right=5pt, top=2pt, bottom=2pt, title=Teacher loss]
\begin{equation}
\label{eq:teacherloss}
\underbrace{\gL_{\mathrm{CE}}\!\left(\mathcal{X};\, \vth_t^T\right)}_{\text{\textbf{Task loss}}} \, + \, \beta^{T}\underbrace{\gL_{\mathrm{drift}}\!\left(\vth_t^T\right)}_{\text{\textbf{Drift \cref{eq:regularization_objective}}}} \,
\end{equation}
\end{tcolorbox}
where \(\mathcal{X} := \{(\vx_i, f_t^\star(\vx_i))\}_{i=1}^B\) denotes a batch of $B$ examples sampled from task \(t\) and $\beta^T$ is a hyperparameter.

\nicepar{Curvature-Aware weight disentanglement.} As discussed in \cref{sec:background}, prior work~\cite{porrello2025dataless} minimizes representation drift~\cite{yoshidamastering} via a KFAC approximation of the generalized Gauss-Newton (GGN) matrix. This strategy, however, requires access \emph{during training} to the KFAC factors of all other tasks to be merged. As a consequence, it assumes that the full set of tasks is known \emph{a priori}, which constitutes a severe limitation in realistic settings. Moreover, when a new task is introduced after training, all previously learned task vectors must be retrained to restore disentanglement with respect to the newly introduced task.

To overcome this limitation, we aim to promote disentanglement in a more task-agnostic manner, producing task vectors that are disentangled with respect to \emph{any} potential input distribution rather than a fixed and predefined set of tasks. To this end, we hypothesize that disentanglement can be achieved by regularizing on a proxy dataset that is sufficiently large and diverse to approximate the underlying data manifold. Concretely, we pre-compute a single, shared curvature matrix on a \emph{\textbf{reference dataset}} \(\mathcal{D}_{\Omega}\). This dataset is intended to capture a broad range of input distributions spanning many possible downstream tasks. Specifically, for vision tasks we estimate curvature on ImageNet-21k~\citep{deng2009imagenet,ridnik1imagenet}, using a randomly sampled 15\% subset of the original training set. For textual tasks, we instead rely on the C4 corpus~\cite{raffel2020exploring}, employing $10^5$ randomly sampled examples\footnote{See \cref{app:ref_data} in the appendix for an ablation on the sensitivity to the choice of the reference dataset $\mathcal{D}_\Omega$.}.

By regularizing each task vector against the curvature matrix derived from the reference dataset \(\mathcal{D}_{\Omega}\), we encourage updates that avoid parameter directions likely to be relevant for other tasks. To obtain a precise estimate of the curvature induced by \(\mathcal{D}_{\Omega}\), we adopt the Eigenvalue-Corrected Kronecker-Factored Approximate Curvature (EK-FAC)~\cite{george2018fast}, which provides a more accurate approximation of the GGN than standard KFAC. While KFAC approximates the GGN as a Kronecker product of two second-moment matrices, \(\mA^l \otimes \mG^l\), EK-FAC further refines this approximation by explicitly modeling the eigenvalues of the curvature in the Kronecker-factored eigenbasis.

With EK-FAC factors pre-computed on $\mathcal{D}_{\Omega}$, the representation drift for any task vector \(\vt\) can be minimized via
\begin{equation}
\label{eq:regularization_objective}
\mathcal{L}_{\mathrm{drift}}(\vth_t)
=
\sum_{l=1}^L
{(\vth_t^{l} - \vth_0^{l})}^\top
\mathrm{GGN}^{l}_{\operatorname{EK\text{-}FAC}}
{(\vth_t^{l} - \vth_0^{l})},
\end{equation}
where \(\mathrm{GGN}^{l}_{\operatorname{EK\text{-}FAC}}
=
(U_A^l \otimes U_G^l)\, S^l\, (U_A^l \otimes U_G^l)^\top\) denotes the EK-FAC approximation of the GGN; here, \(U_A^l\) and \(U_G^l\) are the Kronecker-factored eigenbases and \(S^l\) the diagonal matrix of corrected eigenvalues.

\subsection{Training the Non-Linear Student}
\label{sec:student}
We train the student model $f(\vx;\vth_t^S)$ in the conventional non-linear fine-tuning regime. As detailed in \cref{sec:setup}, transfer is performed in an online fashion, with the two models learning simultaneously. The resulting student loss is:
\begin{tcolorbox}[colback=blue!5!gray!10,boxsep=1pt, left=1pt, right=5pt, top=-4pt, bottom=5pt, title=Student loss]
\begin{equation}
\label{eq:studentloss}
\underbrace{\gL_{\mathrm{CE}}\!\left(\mathcal{X};\,\vth_t^S\right)}_{\text{\textbf{Task loss}}} + \, \beta^{S}\underbrace{\gL_{\mathrm{drift}}\!\left(\vth_t^S\right)}_{\text{\textbf{Drift \cref{eq:regularization_objective}}}}
\, + \, \gamma\underbrace{\gL_{\mathrm{KD}}\!\left(\mathcal{X};\vth_t^S\right)}_{\text{\textbf{Transf.  \cref{eq:transferabilitystudent}}}}
\end{equation}
\end{tcolorbox}
The last term corresponds to a tailored modification of the standard MSE-based distillation objective, which transfers intermediate representations along the teacher’s trajectory (see next paragraph). Finally, we emphasize that feature-level distillation constrains the student to operate close to a linear regime. In this setting, curvature-aware regularization is well defined; accordingly, we incorporate an EK-FAC-based penalty into the student loss in \cref{eq:studentloss}, \ie\ $\gL_{\mathrm{drift}}\left(\vth_t^S\right)$, computed from the GGN on the reference dataset $\mathcal{D}_{\Omega}$, to further promote weight disentanglement.

\nicepar{Along-Path Knowledge Distillation.} Rather than distilling a single teacher model $f_{\mathrm{lin}}(\vx;\vth_t^T)$, we further exploit its linear structure and perform distillation over a continuum of teacher models, each obtained by interpolating the task vector along the linear path originating from the origin $\vth_0$. The Along-Path Knowledge Distillation (\textbf{APKD}) loss is:
\begin{align}
\label{eq:transferabilitystudent}
        \gL_{\mathrm{KD}} (\mathcal{X}; \vth_t^S)
        = &
        \, \mathbb{E}_{\alpha \sim \mathcal{U}(0.5, 1)} \Bigg[
        \frac{1}{B}
        \sum_{i=1}^{B}
        \bigl\|
            f(\vx_i;\vth_0 + \alpha\, \vt_t^S)
        \notag \\
        &
            -
            \texttt{SG}\!\left[
                f_{\mathrm{lin}}(\vx_i;\vth_0 + \alpha\, \vt_t^T)
            \right]
        \bigr\|_2^2
        \Bigg]
        \, . 
\end{align}
In practice, we approximate the expectation by sampling a single $\alpha \sim \mathcal{U}(0.5, 1)$ per optimization step and using the corresponding teacher $f_{\mathrm{lin}}(\vx;\vth_0 + \alpha\, \vt_t^T)$. Also, in \cref{eq:transferabilitystudent}, gradients are prevented from propagating through the teacher via the stop-gradient operator $\texttt{SG}[\cdot]$. This along-path distillation objective encourages the student to inherit the inductive bias of the linearized model not at a single point, but along the linear path between $\vth_0$ and $\vth_t^T = \vth_0 + \vt_t^T $, yielding representations robust to rescaling.

\section{Experiments}
\label{sec:exp}
\begin{table*}[t]
\caption{\textbf{Task Addition.} Performance comparison against in-training merging methods. \textit{Abs.} denotes the absolute accuracy, while \textit{Norm.} represents accuracy normalized by the accuracies of individually fine-tuned models.}
\label{tab:task_addition}
\centering
\setlength{\tabcolsep}{6pt}
\begin{tabular}{llcccccccc}
\toprule
\multirow{3}{*}{\textbf{Method}} & \multirow{3}{*}{$\boldsymbol{\alpha}$}
& \multicolumn{4}{c}{\textbf{\tott{8-Vision}}}
& \multicolumn{2}{c}{\textbf{\tott{14-Vision}}}
& \multicolumn{2}{c}{\textbf{\tott{6-NLI}}} \\ 
\cmidrule(lr){3-6} \cmidrule(lr){7-8} \cmidrule(lr){9-10}
 & & \multicolumn{2}{c}{\textbf{ViT-B/32}} 
 & \multicolumn{2}{c}{\textbf{ViT-L/14}}
 & \multicolumn{2}{c}{\textbf{ViT-B/32}}
 & \multicolumn{2}{c}{\textbf{T5-Base}} \\ 
 & & Abs. & Norm. & Abs. & Norm. & Abs. & Norm. & Abs. & Norm. \\ 
\midrule
Pre-trained & ---  & $48.4$ & --- & $65.0$ & --- & $57.8$ & --- & $61.7$ & --- \\
Individual  & ---  & $92.8$ & --- & $95.8$ & --- & $90.2$ & --- & $85.9$ & --- \\

\midrule
\rowcolor{myredlight} 
\multicolumn{10}{l}{\textbf{Linearized models}} \\
\midrule

\multirow{2}{*}{Linear FT \cite{ortiz2023task}} 
 & $1.0$ & $77.4$ & $88.0$ & $88.0$ & $94.8$ & $73.7$ & $83.4$ & $76.0$ & $92.9$ \\
 & Best & $78.9$ & $89.8$ & $88.0$ & $94.8$ & $76.7$ & $87.0$ & $76.4$ & $93.5$ \\ 
\cmidrule{2-10}
 
\multirow{2}{*}{$\vt \mathbf{Jp}$ \cite{yoshidamastering}} 
 & $1.0$ & $85.0$ & $97.4$ & $90.9$ & $98.3$ & $85.3$ & $97.0$ & $82.5$ & $100.0$ \\ 
 & Best & $85.6$ & $98.2$ & $91.1$ & $98.5$ & $85.4$ & $\mathbf{97.1}$ & $\mathbf{82.5}$ & $\mathbf{100.0}$ \\
\cmidrule{2-10}

\multirow{2}{*}{TAK \cite{porrello2025dataless}} 
 & $1.0$ & $86.0$ & $97.7$ & $91.6$ & $99.3$ & $84.3$ & $95.6$ & $79.1$ & $98.4$ \\ 
 & Best & $86.1$ & $97.8$ & $91.6$ & $99.3$ & $84.7$ & $96.0$ & $79.5$ & $98.8$ \\ 

\midrule
\rowcolor{mybluelight} 
\multicolumn{10}{l}{\textbf{Non-Linear models}} \\
\midrule

\multirow{2}{*}{Non-Linear FT \cite{ilharco2022editing}} 
 & $1.0$ & $32.0$ & $32.9$ & $45.3$ & $47.5$ & $15.6$ & $16.6$ & $42.0$ & $49.7$ \\ 
 & Best & $73.5$ & $80.4$ & $84.5$ & $89.7$ & $68.9$ & $76.1$ & $78.2$ & $91.6$ \\ 
\cmidrule{2-10}

\multirow{2}{*}{TaLoS \cite{iurada2025efficient}} 
 & $1.0$ & $53.3$ & $59.7$ & $46.1$ & $50.8$ & $33.5$ & $37.3$ & $61.7$ & $72.4$ \\
 & Best & $77.9$ & $87.7$ & $84.7$ & $91.1$ & $74.9$ & $84.4$ & $76.7$ & $89.8$ \\
\cmidrule{2-10}

\multirow{2}{*}{Attn. Only FT \cite{jin2024fine}} 
 & $1.0$ & $22.5$ & $23.3$ & $66.2$ & $69.7$ & $13.8$ & $15.1$ & $51.6$ & $61.6$ \\
 & Best & $78.2$ & $86.3$ & $88.2$ & $93.8$ & $73.4$ & $81.5$ & $76.7$ & $91.4$ \\
 
\midrule
\rowcolor{gray!15}
 & $1.0$ & $88.3$ & $98.3$ & $92.7$ & $99.5$ & $85.9$ & $95.9$ & $82.3$ & $95.5$ \\
\rowcolor{gray!15}
\multirow{-2}{*}{\textbf{\methname{}} (ours)}
 & Best & $\mathbf{88.3}$ & $\mathbf{98.3}$ & $\mathbf{92.7}$ & $\mathbf{99.5}$ & $\mathbf{86.0}$ & $96.0$ & $82.4$ & $95.5$ \\
\bottomrule
\end{tabular}
\end{table*}

\nicepar{Vision tasks.} We evaluate our method on two multi-task image classification benchmarks. First, we test \methname{ } on the standard \textbf{\tott{8-Vision}} benchmark~\citep{ilharco2022editing}, which comprises eight heterogeneous visual classification tasks. To assess scalability to larger task pools, we further evaluate on the \textbf{\tott{14-Vision}} benchmark~\citep{gargiulo2025task}, which extends the former with six additional vision tasks, substantially increasing task diversity and difficulty.

\nicepar{Language tasks.} Following~\cite{porrello2025dataless}, we evaluate our framework on the \textbf{\tott{6-NLI}} benchmark~\citep{stoica2024model}, which comprises six Natural Language Inference datasets spanning diverse linguistic domains.

\nicepar{Backbones.} For vision experiments, we use two variants of CLIP~\citep{radford2021learning}, employing ViT-B/32 and ViT-L/14 as visual encoders. For each task, we fine-tune the visual encoder while keeping the text encoder frozen. For language experiments, we adopt the T5-base model~\citep{raffel2020exploring} as backbone for all \tott{6-NLI} tasks. Unless otherwise specified, all methods -- both teacher and student in our case -- employ full fine-tuning as the learning strategy.

\nicepar{Metrics.}
Following the original setup of~\citep{ortiz2023task}, we employ absolute and normalized accuracy. We further analyze the role of the rescaling coefficient $\boldsymbol{\alpha}$: \textit{(i)} fixing $\alpha_t = \alpha = 1$ for all tasks, \ie{} plain summation of task vectors, and \textit{(ii)} tuning $\alpha$ on a cross-task validation set.
\subsection{Comparison with the state-of-the-art}
\label{sec:sota}
We compare against a broad set of existing methods, including both \textbf{in-training} approaches that mitigate task interference during optimization and \textbf{post-hoc} strategies that operate solely at merging time (after training).

\nicepar{\textbf{Task Addition.}} Considering in-training approaches, we first compare against \textit{Non-Linear Fine-Tuning}~\citep{ilharco2022editing} and \textit{Linear Fine-Tuning}~\citep{ortiz2023task}, in which task-specific models are optimized independently. Within the class of non-linear methods, we evaluate \textit{TaLoS}~\citep{iurada2025efficient} and \textit{Attention-Only Fine-Tuning}~\citep{jin2024fine}. The former identifies and updates a sparse subset of parameters with low Fisher sensitivity, while the latter fine-tunes only attention layers. Within the class of linearized models, we compare against \textit{$\vt Jp$}~\citep{yoshidamastering}, a data-dependent regularization method that minimizes representation drift. Finally, we consider TAK~\citep{porrello2025dataless}, a curvature-aware regularization method based on KFAC that reformulates the penalty of $\vt Jp$ to avoid reliance on data from other tasks.

In \cref{tab:task_addition}, we compare our method with state-of-the-art in-training approaches for \textbf{task addition}. Notably, in terms of absolute accuracy, \methname{} consistently achieves strong performance, outperforming all competing approaches across settings, while remaining competitive in terms of normalized accuracy. The comparison with \colorbox{mybluelight}{\textbf{non-linear methods}} is particularly revealing: despite our model being fine-tuned in a regime that is, in principle, non-linear, we observe substantial gains over standard non-linear fine-tuning, with improvements of up to approximately $+15$ absolute accuracy on ViT-B/32 in the \tott{8-Vision} setting. This result highlights the impact of distillation from the linear regime, an aspect we will further dissect in the following.

The gap with \colorbox{myredlight}{\textbf{linearized methods}} narrows, yet our approach remains superior. Notably, this advantage comes with improved inference-time efficiency and greater flexibility, as our method does not require access to task-specific data such as \textit{$\vt Jp$}~\citep{yoshidamastering} nor KFAC statistics~\citep{porrello2025dataless} from other tasks.
\begin{table}[t]
    \centering
    \caption{\textbf{Task Negation.} Unlearning performance on target versus control (ImageNet-1K) tasks.}
    \label{tab:task_negation}
    
    \setlength{\tabcolsep}{3pt}
    \resizebox{\linewidth}{!}{%
        \begin{tabular}{lcccccc}
        \toprule
        & \multicolumn{4}{c}{\textbf{\tott{8-Vision}}} 
        & \multicolumn{2}{c}{\textbf{\tott{14-Vision}}}\\
        \cmidrule(lr){2-5} \cmidrule(lr){6-7}
        \multirow{2}{*}{\textbf{Method}} 
        & \multicolumn{2}{c}{\textbf{ViT-B/32}}
        & \multicolumn{2}{c}{\textbf{ViT-L/14}}
        & \multicolumn{2}{c}{\textbf{ViT-B/32}}\\
        & Targ. & Cont. & Targ. & Cont. & Targ. & Cont. \\
        \midrule
        Pre-trained    & $48.4$ & $63.3$ & $65.0$ & $75.5$ & $57.8$ & $63.3$ \\
        \midrule
        \colorbox{myredlight}{\raisebox{0pt}[1.2ex][0.1ex]{Linear FT}}      & $9.3$  & $60.5$ & $7.1$  & $72.1$ & $19.9$ & $60.6$ \\
        \colorbox{myredlight}{\raisebox{0pt}[1.2ex][0.1ex]{$\vt \mathbf{Jp}$}}
& $3.0$  & $60.6$ & $\mathbf{1.8}$  & $74.2$ & $\mathbf{4.5}$  & $60.8$ \\
        \colorbox{myredlight}{\raisebox{0pt}[1.2ex][0.1ex]{TAK}}            & $\mathbf{2.8}$  & $61.5$ & $3.2$  & $73.6$ & $5.6$  & $60.9$ \\
        \midrule
        \colorbox{mybluelight}{\raisebox{0pt}[1.2ex][0.1ex]{Non-Linear FT}}  & $20.4$ & $60.5$ & $18.1$ & $72.3$ & $30.5$ & $60.6$ \\
        \colorbox{mybluelight}{\raisebox{0pt}[1.2ex][0.1ex]{TaLoS}}          & $18.4$ & $61.1$ & $23.0$ & $74.1$ & $27.3$ & $61.1$ \\
        \colorbox{mybluelight}{\raisebox{0pt}[1.2ex][0.1ex]{Attn. Only FT}}  & $16.7$ & $60.8$ & $17.9$ & $73.3$ & $27.2$ & $60.9$ \\
        \midrule
        \rowcolor{gray!15} \rule{0pt}{2ex}
        \textbf{\methname{}} (ours)
                       & $9.6$  & $\mathbf{62.1}$ & $11.7$ & $\mathbf{74.7}$ & $19.1$ & $\mathbf{62.1}$ \\
        \bottomrule
        \end{tabular}%
}
\end{table}

\nicepar{Task Negation.} We adopt the protocol introduced in~\cite{ilharco2022editing} to forget a target task by subtracting its task vector, and evaluate performance on both the target and a general control task. As shown in \cref{tab:task_negation}, on the \tott{8-Vision} benchmark with ViT-B/32, standard non-linear fine-tuning starts from a target accuracy of $20.4\%$ and a control accuracy of $60.5\%$. In contrast, \methname{} reduces the target accuracy to $9.6\%$ while maintaining a high control accuracy of $62.1\%$ -- outperforming both Attn. Only FT and TaLoS. Notably, our method is outperformed by TAK and $\vt \mathbf{Jp}$, which rely on the linear regime, suggesting that this regime retains residual disentanglement that is more effectively exploited in task subtraction than in addition.
\begin{table}[t]
    \centering
    \caption{\textbf{LoRA-based Task Addition.} Results using LoRA re-parameterization with different merging techniques.}
    \label{tab:lora}
    \setlength{\tabcolsep}{3pt}
    \resizebox{\linewidth}{!}{%
            \begin{tabular}{lcccccc}
\toprule
& \multicolumn{4}{c}{\textbf{\tott{8-Vision}}} 
& \multicolumn{2}{c}{\textbf{\tott{14-Vision}}}\\
\cmidrule(lr){2-5} \cmidrule(lr){6-7}
\multirow{2}{*}{\textbf{Method}} 
& \multicolumn{2}{c}{\textbf{ViT-B/32}}
& \multicolumn{2}{c}{\textbf{ViT-L/14}}
& \multicolumn{2}{c}{\textbf{ViT-B/32}}\\
& Abs. & Norm. & Abs. & Norm. & Abs. & Norm.\\
\midrule
Non-Linear FT & $72.6$ & $79.5$ & $85.3$ & $90.5$ & $68.4$  & $75.7$ \\
Linear FT     & $75.9$ & $86.2$ & $87.1$  & $94.0$ & $75.5$  & $85.5$ \\
\midrule
Iso-C      & $70.6$ & $77.9$ & $85.3$  & $90.7$ & $71.9$  & $80.0$ \\
TSV-M  & $76.4$ & $83.9$ & $88.9$  & $94.4$ & $74.3$  & $82.1$ \\

Core + Iso-C & $73.6$ & $81.2$ & $87.4$  & $93.0$ & $72.3$  & $80.0$ \\
Core + TSV-M & $77.9$ & $85.6$ & $89.1$  & $94.6$ & $74.5$  & $82.5$ \\

\midrule

\rowcolor{gray!15} \rule{0pt}{2ex}
\textbf{\methname{}} (ours)
              & $\mathbf{87.5}$ & $\mathbf{97.9}$ & $\mathbf{92.2}$  & $\mathbf{99.1}$ & $\mathbf{85.7}$  & $\mathbf{96.3}$ \\
\bottomrule
\end{tabular}
}

    \setlength{\tabcolsep}{3pt}
\end{table}

\begin{figure}[t]
    \centering
    \includegraphics[width=0.99\linewidth]{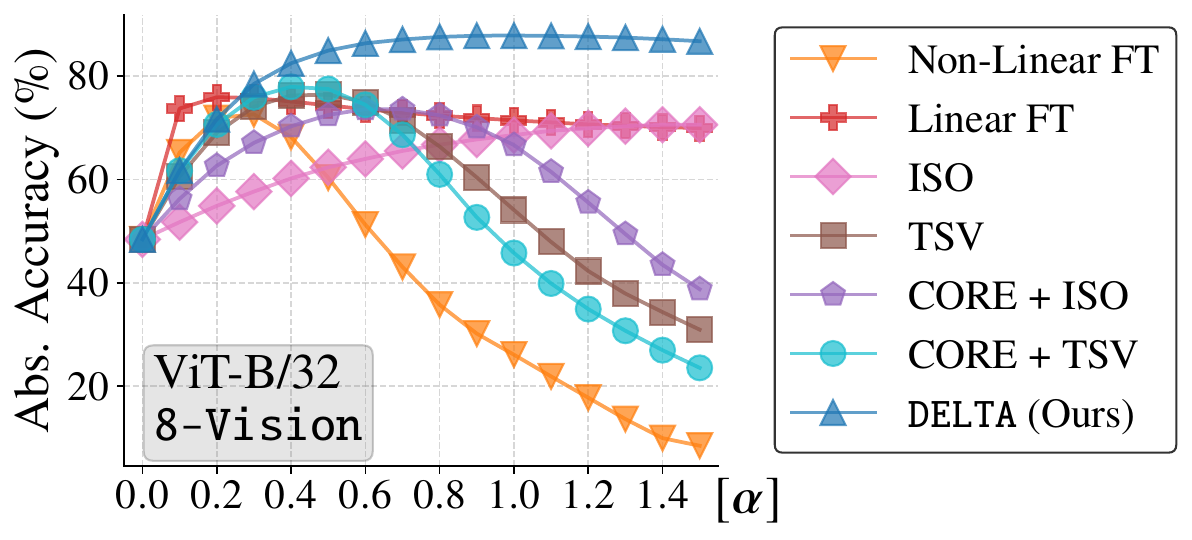}
    \caption{Sensitivity to the scaling coefficient in LoRA merging. Additional results are provided in \cref{fig:lora_sweep_full} of \cref{app:additional_experiments}.}
    \label{fig:lora_sweep}
\end{figure}

\nicepar{LoRA-based Task Addition.} Our approach supports different parameterizations for the teacher and the student. In particular, the teacher can be trained via full fine-tuning, while the student relies on parameter-efficient methods such as LoRA~\cite{hu2022lora}. This design allows the teacher to explore richer task-specific directions in weight space, while constraining the student to a low-rank, efficient subspace.

To further examine this, we compare the \methname{} LoRA student with three state-of-the-art model merging methods: Iso-C~\cite{marczak2025no}, TSV-M~\cite{gargiulo2025task}, and Core Space~\cite{panariello2025accurate}. We remark that these approaches operate \textbf{post hoc}, after standard fine-tuning, and are thus conceptually different from ours, which acts \textbf{during training} and directly produces disentangled task vectors. Nevertheless, this comparison helps position \methname{} relative to widely used merging techniques.

As shown in \cref{tab:lora}, our method outperforms existing approaches by a large margin. Moreover, the $\alpha$-sweep sensitivity analysis in \cref{fig:lora_sweep} shows that, unlike most baselines (except linear fine-tuning), our model is robust to rescaling of the coefficient $\alpha$ used to scale the merged task vector. This is particularly desirable when $\alpha$ cannot be tuned, \eg{} in the absence of a validation set. Taken together, these results suggest that in-training regularization can play a pivotal role in enabling effective fusion of low-rank updates.

\section{Model Analysis}
\begin{figure}[tb]
    \centering
    \includegraphics[width=\linewidth]{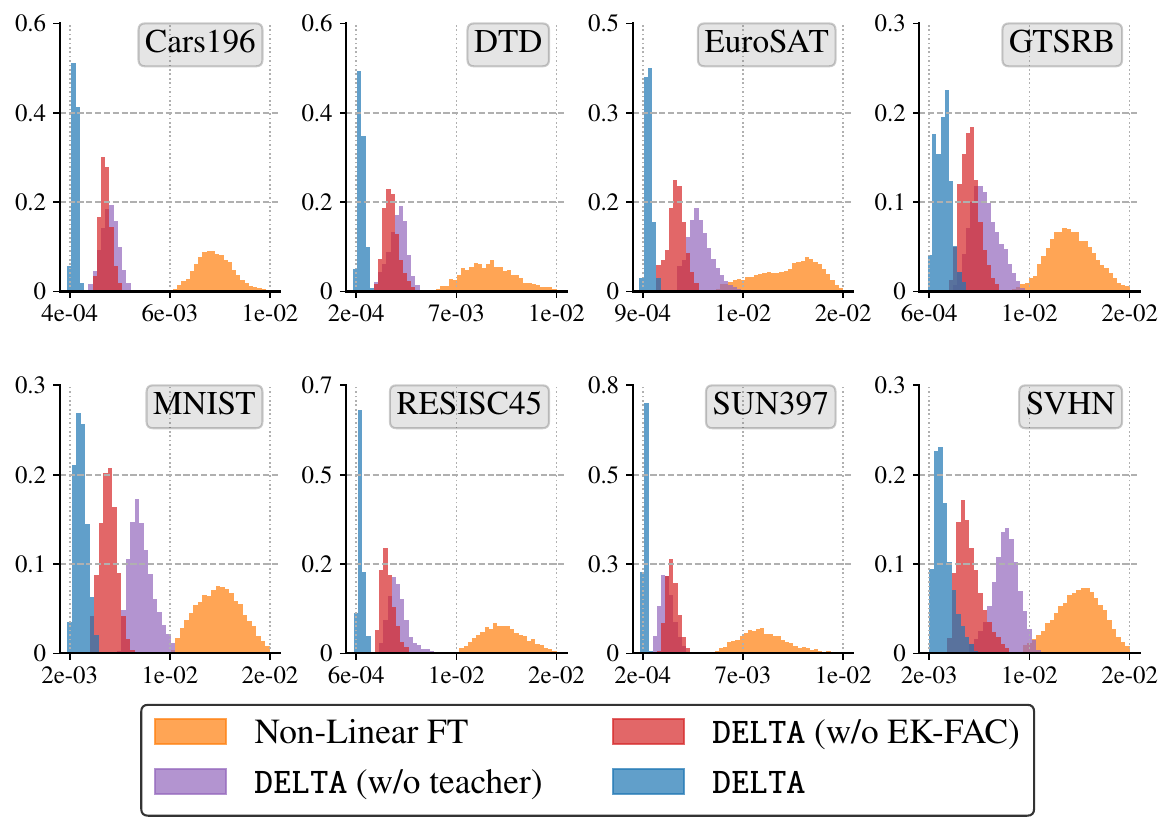}
    \caption{Histogram of the linearization error on \tott{8-Vision}.}
    \label{fig:histo_lin_err}
\end{figure}

\cref{sec:sota} showed that \methname{} outperforms conventional fine-tuning on Task Addition and Negation. Here, we investigate the reasons behind these gains. We focus on two properties learned by the student: the transfer of \textit{linearized behavior} and the emergence of \textit{support localization}. We then assess their relative contributions to task arithmetic performance.

\begin{figure*}[t]
    \centering
    \includegraphics[width=0.95\linewidth]{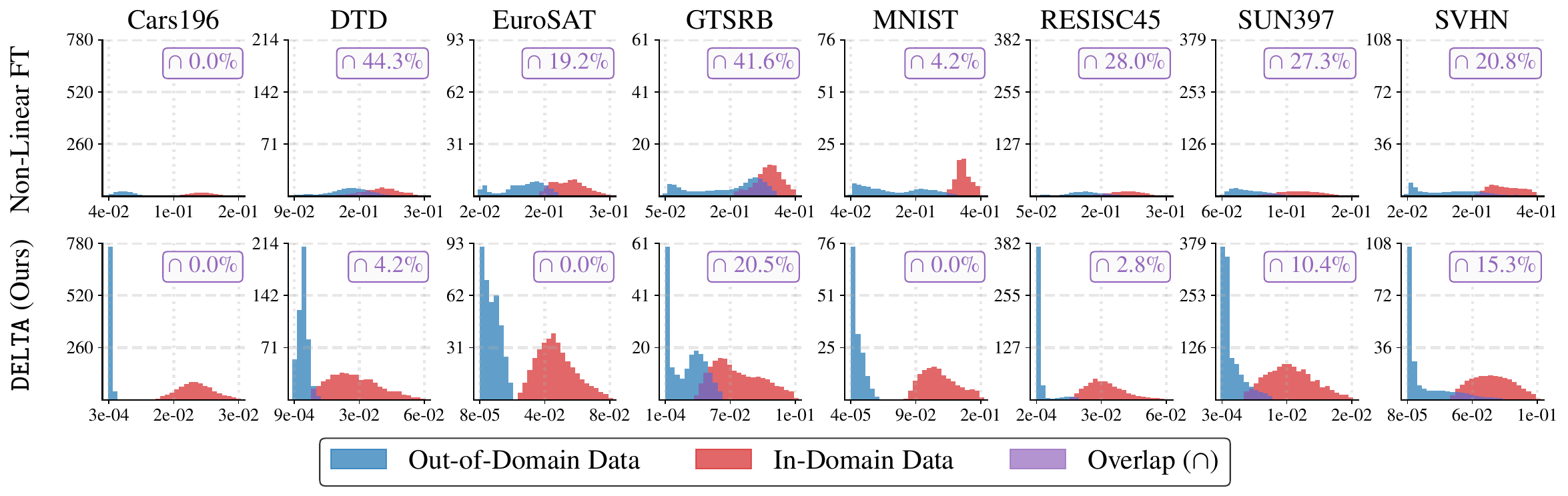}
    \caption{\textbf{Support Localization.} Histograms of the activation MSE between the pre-trained and individually fine-tuned models. In our method, the in-domain and out-of-domain distributions are markedly more separated, indicating stronger weight disentanglement.}
    \label{fig:histo_localization}
\end{figure*}

{\nicepar{Linearization error.} We verify whether the student -- despite operating in a non-linear regime -- exhibits linear behavior under weight perturbations. To this end, we define the \textit{linearization error} on a single example $\vx$ as the discrepancy between the average activation of perturbed models and the activation at the mean weights $\bar{\vth}^S = \frac{1}{T}\sum_{t=1}^T \vth_t^S$:
\begin{equation}
\textstyle
\xi_{\mathrm{lin}}(\vx)=\left\|\frac{1}{T}\sum_{t=1}^T f(\vx; \vth_t^S)-f(\vx; \bar{\vth}^S)\right\|_1.
\end{equation}
This error is exactly zero in the linear regime; thus, a low error indicates an effective transfer of linear behavior. 

In \cref{fig:histo_lin_err}, we report the distribution of this error across each dataset of \tott{8-Vision}. We compare: \textit{(i)} standard non-linear fine-tuning, \textit{(ii)} our student model without teacher distillation, \textit{(iii)} our student model with distillation but without curvature regularization, and \textit{(iv)} our full method \methname{}. The histograms show that our full method achieves \textbf{near-zero linearization error}, substantially reducing the discrepancy relative to the non-linear baseline. Moreover, teacher distillation is the primary driver of this effect: in terms of linearization error, removing distillation leads to a larger degradation than removing curvature regularization.

\nicepar{Support localization.} \citet{ortiz2023task} showed that strict linearity is not required for task arithmetic. Instead, the key condition is \textit{weight disentanglement} (see \cref{sec:background}), whereby a task-specific update primarily affects predictions on its own data distribution while leaving out-of-domain representations largely unchanged. 

To assess support localization, for each model $t$ we measure the \textit{edit distance}, \ie{} the mean squared error (MSE) between pre- and post-fine-tuning activations, $\frac{1}{d}\|f(\vx; \vth_t^S) - f(\vx; \vth_0)\|_2^2$, on both in-domain training examples and out-of-domain examples drawn from other tasks. As shown in \cref{fig:histo_localization}, our student exhibits \textbf{stronger support localization} than conventional fine-tuning: in \methname{}, the edit distance is higher for in-domain examples while remaining near zero on out-of-domain data, indicating a reversion to the pre-trained model for those examples. Importantly, results in \cref{app:additional_experiments} (\cref{fig:histo_mse_suppl}) show that curvature regularization, rather than distillation, is the primary driver of such localization.

With these results, we show that the student converges to regions of parameter space that behave linearly and are task-localized. This behavior is induced by an objective defined in \textit{function space}, which, perhaps surprisingly, leads to tangible effects in \textit{parameter space}. We conjecture that this transfer from function space to parameter space arises because the objective keeps optimization near the pre-trained model, where a first-order Taylor approximation remains accurate. Furthermore, consistent with the simplicity-bias perspective~\cite{huh2024position}, distilling a linearized teacher may drive optimization toward the simplest mechanism that fits this behavior -- \ie{} solutions with approximately linear responses to parameter perturbations.
\begin{figure}[t]
    \centering
    \vspace{-0.6em}
\includegraphics[width=0.95\linewidth]{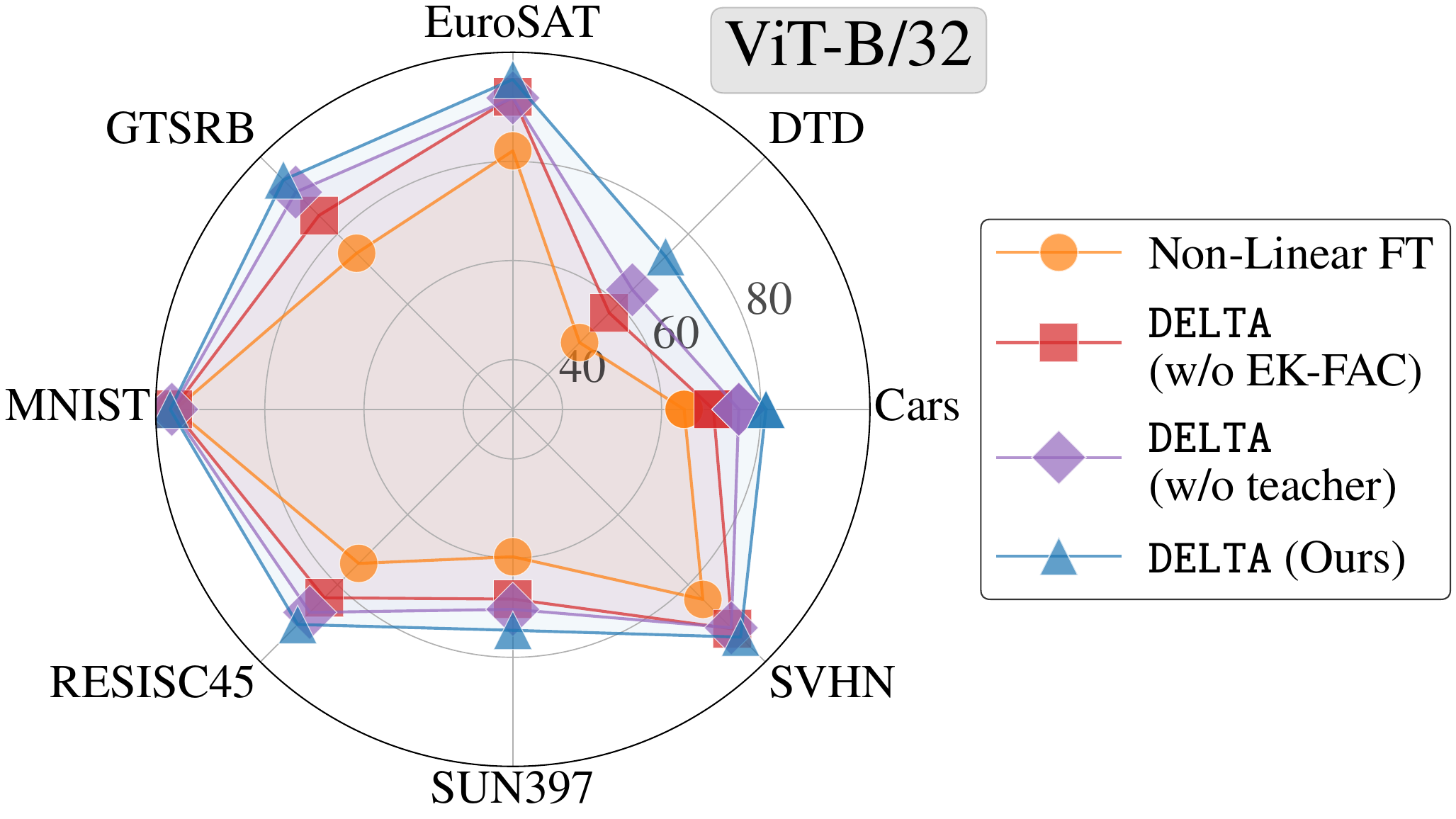}
    \caption{\textbf{Impact on task arithmetic}. Per-task accuracy of the merged model under different configurations. Similar trends persist for ViT-L/14 and T5-Base, as shown in \cref{fig:impact_distillation_full} of \cref{app:additional_experiments}.}
    \label{fig:impact_distillation}
    \vspace{-1.1em}
\end{figure}

\nicepar{Impact on task arithmetic.} We analyze the relative contributions of the two aforementioned properties to task arithmetic. To this end, \cref{fig:impact_distillation} reports per-task performance of the merged model under varying configurations. As shown, distillation from the linear regime alone yields a substantial improvement over conventional fine-tuning. This highlights a possible, more practical variant of our approach that requires no curvature estimation while still delivering meaningful gains. Similarly, applying curvature regularization alone yields performance closest to our full method (though still inferior), confirming that enforcing localized support is the \textbf{most critical condition} for task arithmetic. 

\nicepar{On Along-Path Knowledge Distillation (APKD).} Unlike conventional KD, our method distills knowledge from multiple models (\cref{sec:student}), exposing the student to a continuum of linearized, interpolated teachers, and promoting the transfer of linear dynamics. To validate this, \cref{fig:histo_mse} reports the linearization error when the teacher is fixed at $\alpha=1$ (\ie{} without along-path sampling). As shown, fixing the teacher increases this error across datasets, supporting our claims. Furthermore, we evaluate how teacher sampling influences performance via an $\alpha$-sweep analysis, reporting in \cref{fig:plotalphasweep} the accuracy of the merged model as a function of the rescaling coefficient $\alpha$. Especially on T5, removing teacher sampling leads to significantly less robust performance across $\alpha$ values, highlighting the key role of APKD in transferring the robustness of the linear regime to task vector rescaling.

\begin{figure}
    \centering
    \includegraphics[width=\linewidth,clip,trim=2 2 2 2]{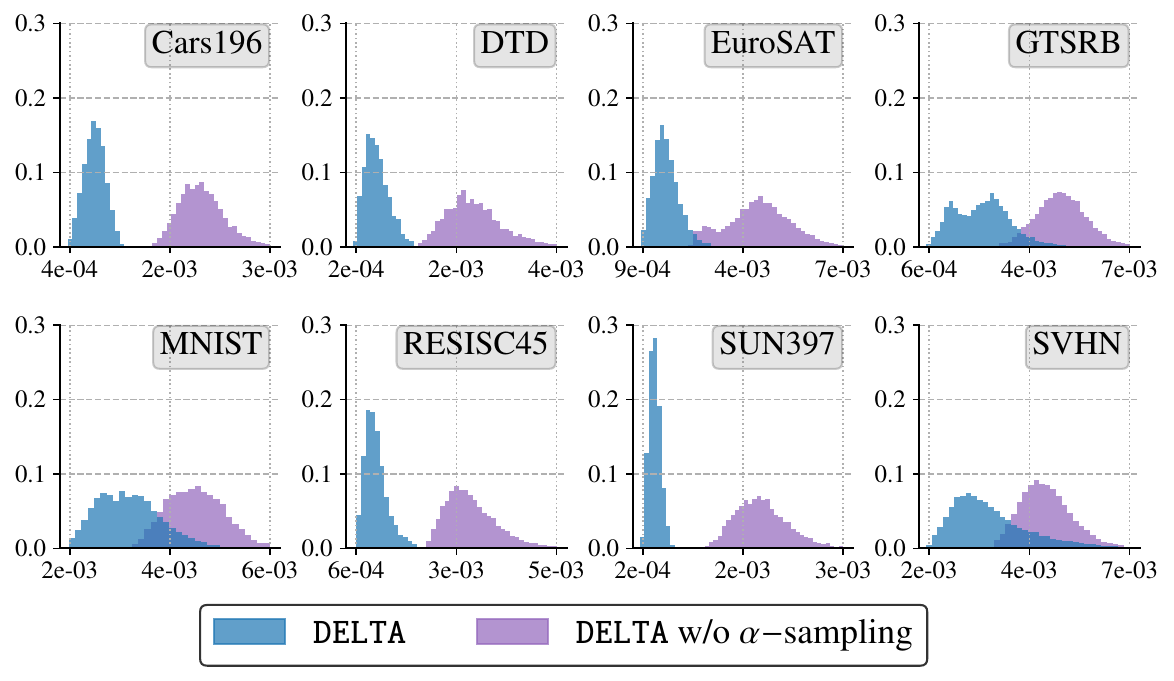}
    \caption{Effect of APKD on linearization error.}
    \label{fig:histo_mse}
\end{figure}

\begin{figure}[t]
    \centering
    \includegraphics[width=\linewidth,clip,trim=2 2 2 2]{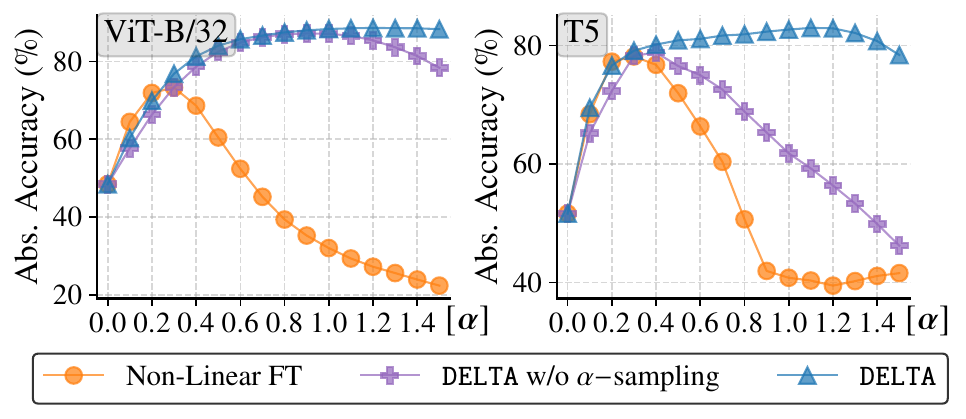}
    \caption{Effect of APKD on robustness to $\alpha$-sweep.}
    \vspace{-0.5em}
    \label{fig:plotalphasweep}
\end{figure}

\begin{figure}[t]
    \centering
    \makebox[\columnwidth][c]{%
        \includegraphics[width=0.49\columnwidth]{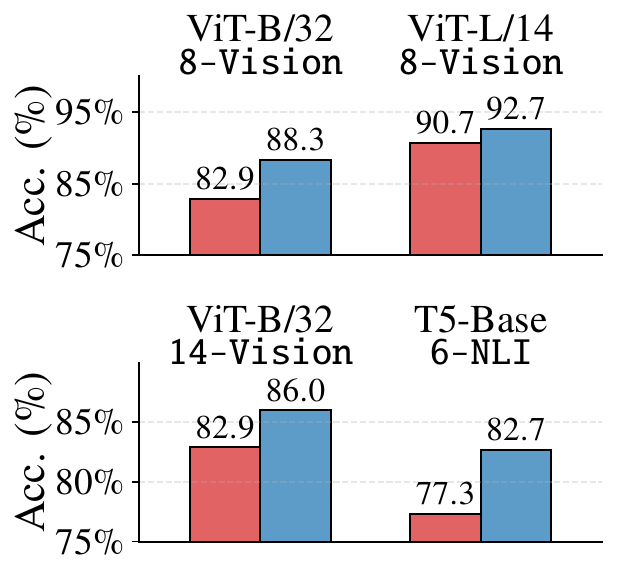}%
        \hspace{0.02\columnwidth}%
        \includegraphics[width=0.49\columnwidth]{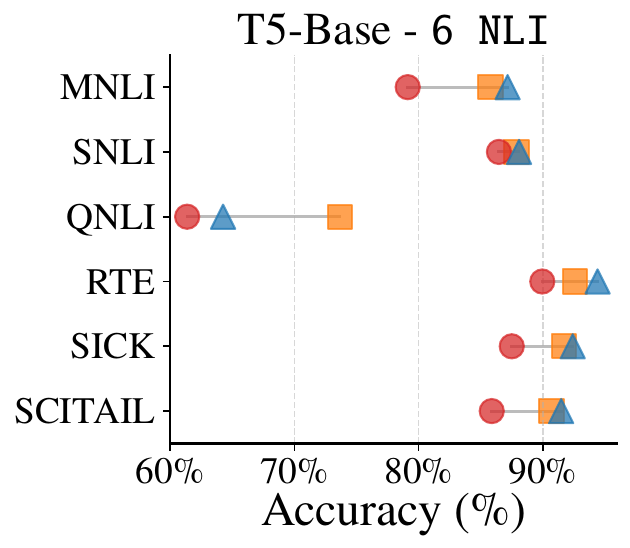}%
    }\\[-0.2em]
    \includegraphics[width=0.98\columnwidth]{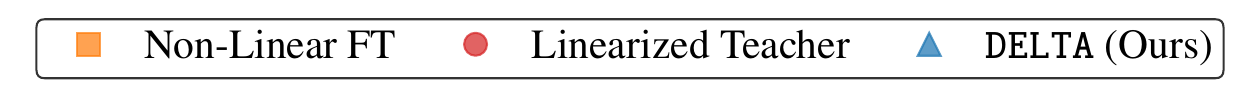}
    \caption{\textbf{Teacher-Student comparison.}
    Left: Accuracy of the merged models.
    Right: Accuracy of the single task fine-tunings.}
    \label{fig:teacher_student_combined}
\end{figure}

\begin{figure}
    \centering
    \includegraphics[width=\linewidth]{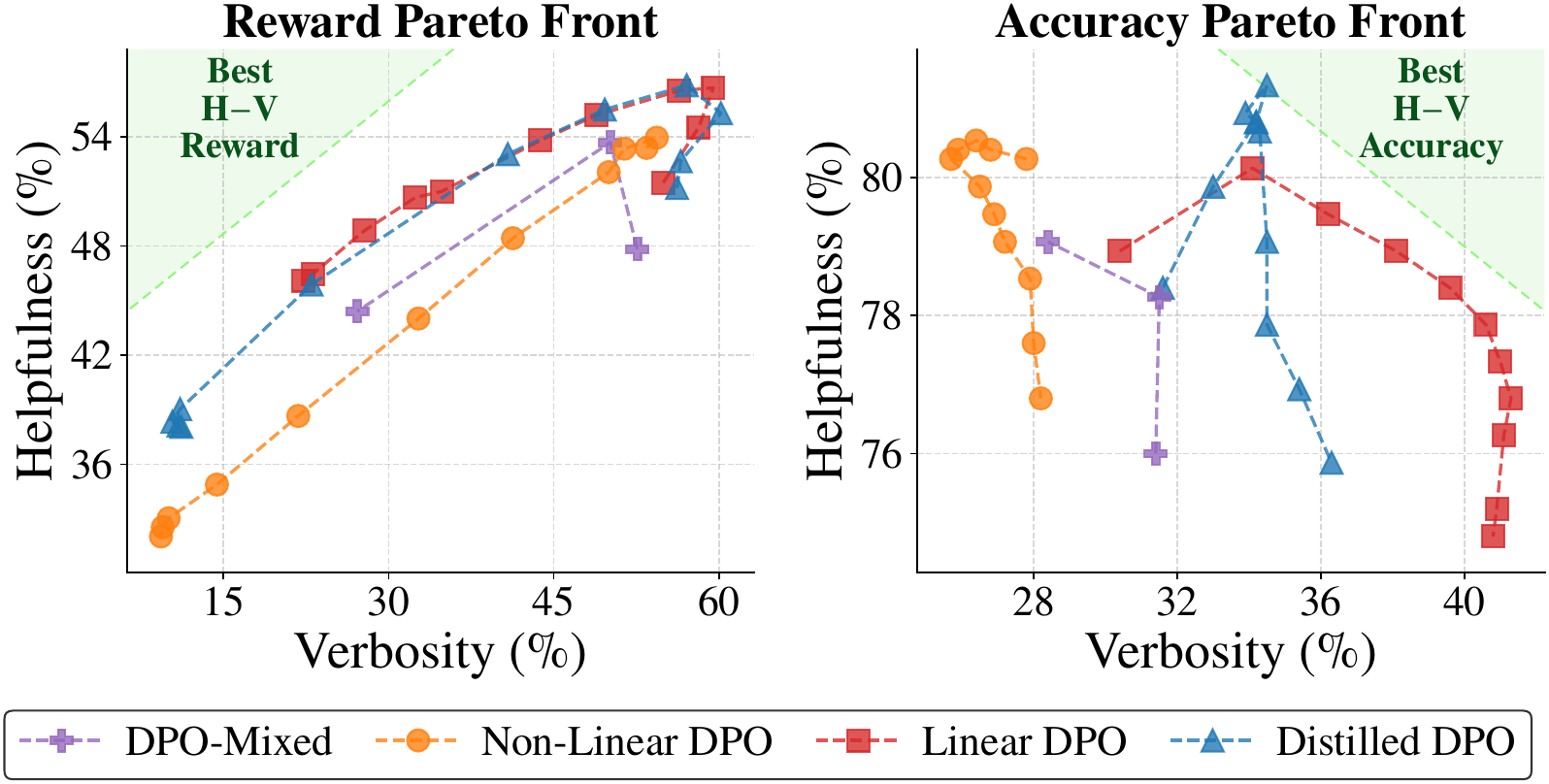}
    \caption{\textbf{Multi-Objective Alignment in Generative LLMs.} Pareto frontiers of the trade-off between helpfulness and verbosity.}
    \label{fig:pareto_llm}
    \vspace{-0.2em}
\end{figure}

\nicepar{On expressivity.} A somewhat surprising finding is that the student consistently surpasses its linearized teacher in average accuracy after merging (see \cref{fig:teacher_student_combined}, \textit{left}). We attribute this to the greater expressivity of the non-linear regime, which allows the student to better fit individual tasks than its teacher. This is corroborated by \cref{fig:teacher_student_combined} (\textit{right}), which reports per-task performance \textbf{before merging} for T5 (additional results are provided in \cref{fig:suppl_teachervsstudent} of \cref{app:additional_experiments}): the student outperforms the linearized teacher on all individual tasks. This suggests that \methname{} does not merely replicate linearized behavior; rather, the student operates in an intermediate, quasi-linear regime that combines the capacity of the non-linear regime with the disentanglement of the linear one.

\nicepar{Extension to Generative LLMs.} Following \citet{erdogan2026tangent}, we go beyond classification and investigate whether task arithmetic can balance multiple preference dimensions in LLMs -- such as \textit{helpfulness} and \textit{verbosity} -- in a controllable manner. Unlike methods such as Direct Preference Optimization (DPO)~\cite{rafailov2023direct}, which collapse multiple axes into a single scalar objective, task arithmetic enables training models specialized along each preference
axis and combining them at inference time. This allows trading off between objectives without further optimization.

We adopt LLaMA-3.2-1B-Instruct~\cite{grattafiori2024llama} and fine-tune two separate models using DPO on UltraFeedback~\cite{cui2024ultrafeedback} and HelpSteer2~\cite{wang2024helpsteer} for helpfulness and verbosity, respectively. The resulting models are merged using the \textit{Affine-2} formulation~\citep{erdogan2026tangent}: $\vth_{\text{mix}}=\vth_{0}+\vt_{\text{help}}+\lambda_2\vt_{\text{verb}}$. We fix the helpfulness direction and sweep the verbosity coefficient $\lambda_2 \in [0,5]$, progressively favoring more concise responses. We assess the mixed models using both reward scores (via a Mistral-7B-based reward model) and pairwise preference accuracy.

We conduct a preliminary study by applying \methname{} to DPO -- called \textit{Distilled DPO} -- using only distillation from the linearized teacher, without curvature regularization. We compare against: \textit{(i)} \textit{DPO-Mixed}, which optimizes a single scalarized objective over the combined dataset; \textit{(ii)} \textit{Non-Linear DPO}, and \textit{(iii)} \textit{Linear DPO}, which combine standard and linear fine-tuning with task arithmetic, respectively. As shown in \cref{fig:pareto_llm}, Distilled DPO closely matches the reward-score Pareto frontier of Linear DPO, achieving a favorable trade-off between helpfulness and verbosity while retaining inference-time efficiency. In terms of preference accuracy, our method surpasses both DPO-Mixed and Non-Linear DPO, although it slightly trails Linear DPO.

\section{Conclusions and Limitations}
We show that linearized behavior can be transferred to ordinary non-linear models, preserving the benefits of standard fine-tuning while improving the composability of task vectors. We describe two complementary effects: distillation induces quasi-linear behavior under weight perturbations, while curvature-aware regularization promotes support localization and reduces task interference. Together, these properties yield models that are easier to compose, tune, and deploy. Preliminary results further suggest that these principles can support controllable multi-objective alignment. The main drawback is the \textbf{training footprint} (\cref{app:computational_costs}): training time increases by a factor of three, while memory usage doubles compared to conventional training. Improving training efficiency is an important direction for future work.

\section*{Acknowledgements}

We acknowledge the CINECA award under the ISCRA initiative, for the availability of high-performance computing resources and support. This work is supported by the Horizon Europe Chips Joint Undertaking under the NexTArc project (HORIZON-JU-Chips-2024-2-RIA). NexTArc -- Next Generation Open Innovations in Trustworthy Embedded AI Architectures for Smart Cities, Mobility and Logistics (Grant Agreement ID: 101194287, DOI: 10.3030/101194287). Additionally, the research activities of Angelo Porrello have been partially supported by the Department of Engineering ``Enzo Ferrari'' through the program FAR2025DIP (CUP E93C25000370005). We also gratefully acknowledge Symboolic s.r.l. for funding the PhD position of Thomas Sommariva. Finally, special thanks to Pietro Buzzega for the many valuable discussions, thoughtful suggestions, and insightful feedback throughout this work.

\section*{Impact Statement}
Model merging is often treated as a purely post-hoc operation applied to independently fine-tuned checkpoints. In this view, users may download models from public repositories, such as Hugging Face, without having controlled their fine-tuning process. This work contributes instead to a \textbf{training-aware view} of model merging, which applies to scenarios where the same entity controls both fine-tuning and merging. For instance, an organization may train a library of composable-by-design models specialized along different behavioral axes and later combine them to provide personalized models without further optimization. Similarly, merging can be used for rehearsal-free continual learning, where new capabilities can be incrementally added to a base model by composing task vectors. In such settings, where the entity performing merging also controls the training process, restricting model merging to post-hoc procedures is unnecessarily limiting: one can instead introduce in-training regularization objectives that explicitly encourage learned updates to be more composable.

In broader terms, this work may support the democratization of AI by making task-specific model updates more modular, reusable, and composable, thereby enabling rapid, low-cost adaptation. Moreover, controllable composition of task vectors may support \textbf{more pluralistic AI systems}, where different capabilities, preferences, or behavioral axes can be combined to better reflect diverse user and application requirements. Furthermore, by improving task subtraction over standard non-linear fine-tuning, \methname{} may facilitate \textit{machine unlearning}. This is a relevant mechanism for complying with data privacy regulations (\eg{} the ``right to be forgotten''), mitigating systemic biases, and efficiently removing toxic concepts from deployed models. Finally, by preserving the inference-time efficiency of standard non-linear fine-tuning, \methname{} may make task arithmetic more practical in resource-constrained settings.

More effective model editing and merging can be a \textbf{double-edged sword}. The same mechanisms that allow benign customization could also be used to combine unsafe capabilities or weaken safety-relevant behaviors, increasing the risk of proliferating harmful AI systems. If task vectors encode undesirable, biased, or unsafe behaviors, improved composability may make it easier to transfer or combine such behaviors across models. In multi-objective alignment settings, composing preference directions could provide useful control, but may also produce unintended trade-offs if the objectives are misspecified or insufficiently evaluated.

Moreover, \methname{} introduces additional training-time cost due to teacher--student optimization and curvature pre-computation, with associated computational and environmental considerations. On the other hand, this cost is incurred once during training, while the resulting student has the same inference-time efficiency as a standard non-linear model. Whether this trade-off is beneficial depends on the deployment scenario: it is most favorable when a merged model is reused many times or when avoiding linearized inference substantially reduces serving cost.

\bibliography{references}
\bibliographystyle{icml2026}

\newpage
\appendix
\onecolumn

\section*{Appendix}
The appendix is organized as follows:
\begin{itemize}
    \item \Cref{app:computational_costs} provides an analysis of the computational footprint, including memory requirements, training overhead, and inference costs.
    \item \Cref{app:implementation_details} details the implementation of our methods, with separate discussions for the vision and text domains.
    \item \Cref{app:ref_data} presents an ablation study investigating the robustness of our method to the choice of the reference dataset $\mathcal{D}_\Omega$.
    \item \Cref{app:additional_experiments} presents additional experiments and extended evaluations.
\end{itemize}

\section{Computational Costs}
\label{app:computational_costs}
\begin{table*}[hb]
\centering
\caption{Comparison of training and inference costs across different methods and architectures. Training time is measured in ms/iteration (forward + backward). Note that at inference, our method employs the non-linear student architecture, matching its efficiency.}
\label{tab:unified_costs}
\begin{tabular}{l cccccccc}
\toprule
 & \multicolumn{4}{c}{\textbf{Training}} & \multicolumn{4}{c}{\textbf{Inference}} \\
\cmidrule(lr){2-5} \cmidrule(lr){6-9}
 & \multicolumn{2}{c}{\textbf{ViT-B/32}} & \multicolumn{2}{c}{\textbf{ViT-L/14}} & \multicolumn{2}{c}{\textbf{ViT-B/32}} & \multicolumn{2}{c}{\textbf{ViT-L/14}} \\
\cmidrule(lr){2-3} \cmidrule(lr){4-5} \cmidrule(lr){6-7} \cmidrule(lr){8-9}
\textbf{Method} & \makecell{Mem \\ (GB)} & \makecell{Time \\ (ms)} & \makecell{Mem \\ (GB)} & \makecell{Time \\ (ms)} & \makecell{Time \\ (ms)} & \makecell{Mem \\ (GB)} & \makecell{Time \\ (ms)} & \makecell{Mem \\ (GB)} \\
\midrule
Linear FT & $10.8$ & $497$ & $34.3$ & $1914$ & $320$ & $3.33$ & $1152$ & $9.36$ \\
Non-Linear FT & $7.7$ & $223$ & $26.2$ & $988$ & $160$ & $3.03$ & $364$ & $8.27$ \\
\midrule
\rowcolor{gray!15}
\methname{} (ours) & $15.3$ & $876$ & $54.4$ & $3562$ & $160$ & $3.03$ & $364$ & $8.27$ \\
\bottomrule
\end{tabular}
\end{table*}

In this section, we detail the memory footprint, training time, and inference costs associated with our approach.

\nicepar{Training overhead and memory footprint.} The training process involves the joint optimization of both the student and teacher models. In~\cref{tab:unified_costs}, we report the training speeds. Our approach requires forward and backward passes through both models, which are executed concurrently. As a result, the training time is approximately three times that of the non-linear fine-tuning regime, representing a practical limitation of our approach.

Regarding peak GPU memory usage, our approach requires approximately $1.5$--$2\times$ the memory of standard linear fine-tuning and approximately $2\times$ that of standard non-linear fine-tuning, as detailed in~\cref{tab:unified_costs}.

It is important to note that these measurements reflect a baseline implementation without any specific memory optimizations. The training footprint can be substantially mitigated in practice: since the backward passes of the student and teacher models are independent, either model can be offloaded from the GPU. Additional savings can be achieved by restricting fine-tuning to a subset of layers or by integrating parameter-efficient fine-tuning (PEFT) strategies, such as LoRA. Moreover, the distillation and curvature-regularization terms could be applied only periodically, rather than at every optimization step, or activated only after an initial phase of fine-tuning, further reducing training time.

\nicepar{Inference costs and deployment efficiency.}
The increased training requirements represent a \textit{one-time cost} in the lifecycle of the resulting models. At deployment time, this cost is amortized by the efficiency of our merging pipeline. 

As reported in~\cref{tab:unified_costs}, the distilled non-linear student model achieves significant latency improvements during a forward pass compared to the linearized teacher model. Measured on a single A100 GPU, the student model yields a $2\times$ speedup on ViT-B/32 and a $3.2\times$ speedup on ViT-L/14, alongside a reduction in peak memory usage. The computational overhead of the linearized model primarily stems from the computationally expensive Jacobian-vector products required during inference, which our distilled student model circumvents entirely. 

Ultimately, our approach enables highly scalable model composition via simple task arithmetic. It completely bypasses the need for complex, post-hoc procedures (\eg{} SVD-based methods) or expensive linearization operations at deployment. This efficiency is particularly advantageous when merging large-scale models, where traditional SVD methods become prohibitively expensive and must be repeatedly re-computed for different weight configurations. By resolving these bottlenecks, our framework effectively supports dynamic settings -- such as pluralistic alignment -- where model compositions must be efficiently adjusted on-the-fly at inference time to match varying user preferences.
\section{Implementation Details}
\label{app:implementation_details}
This section details the datasets and implementation specifics used in the experiments presented in the paper. The overall training procedure is summarized in~\cref{alg:teacher_student_training}. The code for replicating our method is available at \url{https://github.com/apanariello4/merge-and-rebase}~\cite{panariello2026mergeandrebase}.

\nicepar{Vision domain.}
We evaluate our framework on the \textbf{\tott{8-Vision}} benchmark~\citep{ilharco2022editing} which comprises eight heterogeneous image classification datasets: Stanford Cars~\citep{krause20133d}, DTD~\citep{cimpoi2014describing}, EuroSAT~\citep{helber2019eurosat}, GTSRB~\citep{stallkamp2011german}, MNIST~\citep{lecun2002gradient}, RESISC45~\citep{cheng2017remote}, SUN397~\citep{xiao2016sun}, and SVHN~\citep{netzer2011reading}. 
The \textbf{\tott{14-Vision}} benchmark~\citep{gargiulo2025task} serves as an extension for scalability analysis, incorporating six additional classification datasets: CIFAR100~\citep{krizhevsky2009learning}, STL10~\citep{coates2011analysis}, Flowers102~\citep{nilsback2008automated}, Oxford-IIIT Pet~\citep{parkhi2012cats}, PCAM~\citep{veeling2018rotation}, and FER2013~\citep{goodfellow2013challenges}.

For training vision task vectors, we followed the setup of previous works~\cite{ilharco2022editing,ortiz2023task,yoshidamastering}, adopting a batch size of $128$. We used the AdamW optimizer~\cite{loshchilovdecoupled} with a learning rate of $1 \times 10^{-5}$, weight decay of $0.1$, and a cosine annealing learning rate scheduler. For TAK~\citep{porrello2025dataless}, we follow the original paper and use a learning rate of $3 \times 10^{-4}$. Unlike prior approaches, we did not apply gradient clipping during training.   
We weight the regularization terms in the losses \cref{eq:teacherloss} and \cref{eq:studentloss} by \changed{$\beta^T=\beta^S = 500$}. The distillation loss term in \cref{eq:studentloss} is weighted by $\gamma=1.0$. We grid the coefficient $\alpha$ in $(0;1]$ for task addition and in $(0;2]$ for task negation.

\nicepar{Language domain.}
Natural Language Inference (NLI) experiments are tested on the \textbf{\tott{6-NLI}} benchmark \citep{stoica2024model}, which includes six datasets: SNLI~\citep{bowman2015snli}, MultiNLI~\citep{williams2017broad}, SICK~\citep{marelli2014semeval}, which are three-way classification tasks where the relation between a premise and a hypothesis must be identified as entailment, contradiction, or neutral. In contrast, SciTail~\citep{khot2018scitail}, RTE~\citep{wang2018glue}, and QNLI~\citep{wang2018glue} are binary entailment tasks, and therefore fine-tuning and evaluation are restricted to two labels. 

For training language task vectors, we adopted a batch size of $128$, using an AdamW optimizer~\cite{loshchilovdecoupled} with a learning rate of $3\times 10^{-4}$, an iteration-based cosine-annealing scheduler and a weight decay of $0.01$. Like in vision tasks, we did not apply gradient clipping during training. The regularization term in the losses \cref{eq:teacherloss} and \cref{eq:studentloss} is set to $\beta^T=\beta^S =20$ and the distillation loss term in \cref{eq:studentloss} is weighted by $\gamma=0.01$.

\begin{algorithm}[t]
  \caption{{\methnamecomplete{}} ({\methname{}})}
  \label{alg:teacher_student_training}
  \begin{algorithmic}
    \STATE {\bfseries Input:} Pre-trained initialization $\boldsymbol{\theta}_0$, target task $t$, reference dataset $\mathcal{D}_{\Omega}$, hyper-parameters $\beta^T, \beta^S, \gamma$, $\eta$
    
    \STATE \textbf{Pre-computation:} Compute EK-FAC curvature matrices (GGN factors $U_A^l, U_G^l, S^l$) on the reference dataset $\mathcal{D}_{\Omega}$
    
    \STATE \textbf{Initialize} linearized teacher $\boldsymbol{\theta}_t^T \leftarrow \boldsymbol{\theta}_0$ and non-linear student $\boldsymbol{\theta}_t^S \leftarrow \boldsymbol{\theta}_0$
    \STATE \textbf{Define} task vectors: $\boldsymbol{\tau}_t^T \leftarrow \boldsymbol{\theta}_t^T - \boldsymbol{\theta}_0$ and $\boldsymbol{\tau}_t^S \leftarrow \boldsymbol{\theta}_t^S - \boldsymbol{\theta}_0$
    
    \WHILE{training not converged}
      \STATE Sample a batch $\mathcal{X} = \{(\boldsymbol{x}_i, y_i)\}_{i=1}^B$ from task $t$
      
      \STATE Sample interpolation scalar $\alpha \sim \mathcal{U}(0.5, 1)$
      
      \STATE \quad $\mathcal{L}_{\mathrm{KD}}^S \leftarrow \frac{1}{B} \sum_{i=1}^{B} \left\| f(\boldsymbol{x}_i; \boldsymbol{\theta}_0 + \alpha\boldsymbol{\tau}_t^S) - \texttt{SG}\!\left[f_{\mathrm{lin}}(\boldsymbol{x}_i; \boldsymbol{\theta}_0 + \alpha\boldsymbol{\tau}_t^T)\right] \right\|_2^2$
      
      \STATE \quad $\mathcal{L}^T \leftarrow \mathcal{L}_{\mathrm{CE}}(\mathcal{X}; \boldsymbol{\theta}_t^T) + \beta^T \mathcal{L}_{\mathrm{drift}}(\boldsymbol{\theta}_t^T)$
      \STATE \quad $\mathcal{L}^S \leftarrow \mathcal{L}_{\mathrm{CE}}(\mathcal{X}; \boldsymbol{\theta}_t^S) + \beta^S \mathcal{L}_{\mathrm{drift}}(\boldsymbol{\theta}_t^S) + \gamma \mathcal{L}_{\mathrm{KD}}(\mathcal{X}; \boldsymbol{\theta}_t^S)$
      
    \STATE \quad $\boldsymbol{\theta}_t^T \leftarrow \boldsymbol{\theta}_t^T - \eta \nabla_{\boldsymbol{\theta}_t^T} \mathcal{L}^T$
    \STATE \quad $\boldsymbol{\theta}_t^S \leftarrow \boldsymbol{\theta}_t^S - \eta \nabla_{\boldsymbol{\theta}_t^S} \mathcal{L}^S$      
    \ENDWHILE
\end{algorithmic}
\end{algorithm}

\section{Ablation of the reference dataset $\mathcal{D}_\Omega$}
\label{app:ref_data}
\begin{table*}
\caption{\textbf{Robustness to the choice of $\mathbf{\mathcal{D}_\Omega}$.} Performance comparison across Vision and Text modalities when different reference datasets are used for our regularization technique. \textit{Abs.} denotes the absolute accuracy, while \textit{Norm.} represents accuracy normalized by the accuracies of individually fine-tuned models.}
\label{tab:dataset_performance}
\centering
\setlength{\tabcolsep}{8pt}
\begin{minipage}[t]{0.50\textwidth}

\centering
\begin{tabular}{llcc}
\toprule
\multirow{2}{*}{Method} & \multirow{2}{*}{$\boldsymbol{\alpha}$} 
& \multicolumn{2}{c}{\textbf{Accuracy}} \\ 
\cmidrule{3-4}
& & Abs. & Norm. \\ 
\midrule
\textbf{Vision (ViT-B/32)}& & & \\
\midrule
\multirow{2}{*}{Non-Linear FT} 
 & $1.0$  & $32.0$ & $32.9$ \\
 & Best & $73.5$ & $ 80.4$  \\
\cmidrule{2-4}
\textbf{\methname{}} & $1.0$ & $88.3$ & $98.3$ \\
\quad Reg. ImageNet-21k (Ours) & Best  & $88.3$ & $98.3$ \\
\cmidrule{2-4}
\textbf{\methname{}} & $1.0$ & $87.5$ & $96.0$ \\
\quad Reg. ImageNet-1k & Best  & $88.0$ & $97.3$ \\ 
\bottomrule
\end{tabular}
\end{minipage}\hfill
\begin{minipage}[t]{0.46\textwidth}
\centering
\begin{tabular}{llcc}
\toprule
\multirow{2}{*}{Method} & \multirow{2}{*}{$\boldsymbol{\alpha}$} 
& \multicolumn{2}{c}{\textbf{Accuracy}} \\ 
\cmidrule{3-4}
& & Abs. & Norm. \\ 
\midrule
\textbf{Text (T5-Base)}& & & \\
\midrule
\multirow{2}{*}{Non-Linear FT } 
 & $1.0$  & $42.0 $ & $49.7 $ \\ 
 & Best &  $78.2 $ & $91.6 $ \\ 
\cmidrule{2-4}
\textbf{\methname{}} & $1.0$ & $82.3$ & $95.5$ \\
\quad Reg. C4 (Ours) & Best  & $82.4$ & $95.5$ \\ 
\cmidrule{2-4}
\textbf{\methname{}} & $1.0$ & $80.3$ & $93.4$ \\
\quad Reg. PAWS & Best  & $81.3$ & $92.3$ \\
\cmidrule{2-4}
\textbf{\methname{}} & $1.0$ & $81.6$ & $93.4$ \\
\quad Reg. QQP & Best  & $82.0$ & $94.0$ \\
\bottomrule
\end{tabular}
\end{minipage}

\end{table*}

To investigate the robustness to the choice of the reference data $\mathcal{D}_\Omega$, \cref{tab:dataset_performance} evaluates how varying the dataset used for curvature estimation affects performance. Specifically, we substitute ImageNet-21k with the smaller ImageNet-1k~\cite{deng2009imagenet} for vision tasks, and replace the broad C4~\cite{raffel2020exploring} corpus with smaller datasets, such as PAWS~\cite{zhang2019paws} and QQP~\cite{qqp}, for language tasks. As shown, while the choice of reference dataset has a measurable impact, the overall performance remains highly robust. In all cases, the student model maintains state-of-the-art results, outperforming all competing methods from \cref{tab:task_addition} across nearly all evaluation settings.

\section{Additional experiments.}
\label{app:additional_experiments}
\nicepar{Weight Disentanglement.} In \cref{fig:supp_tareg}, we extend our evaluation and include Attention-Only Fine-Tuning~\citep{jin2024fine}, as well as Linear Fine-Tuning~\citep{ortiz2023task}. This comparison demonstrates that \methname{} achieves a significantly lower disentanglement error compared to all other evaluated fine-tuning techniques.

\begin{figure}
  \centering
  \includegraphics[width=0.82\linewidth,clip,trim=0 18 0 0]{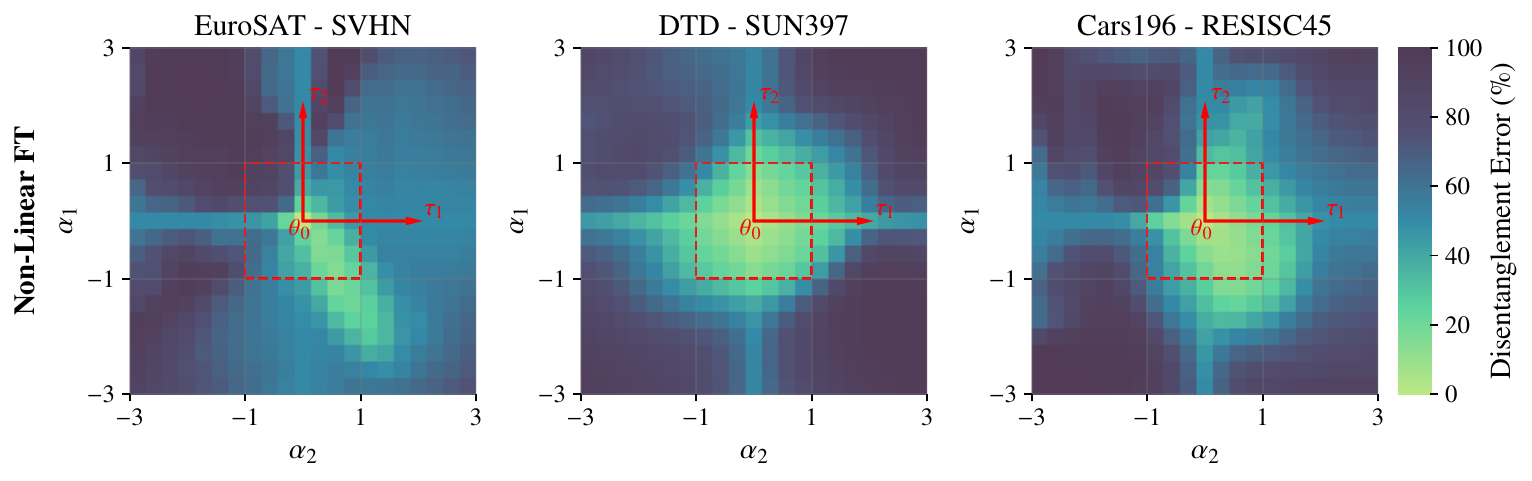}
  \vspace{-1em}
  \includegraphics[width=0.82\linewidth,clip,trim=0 0 0 18]{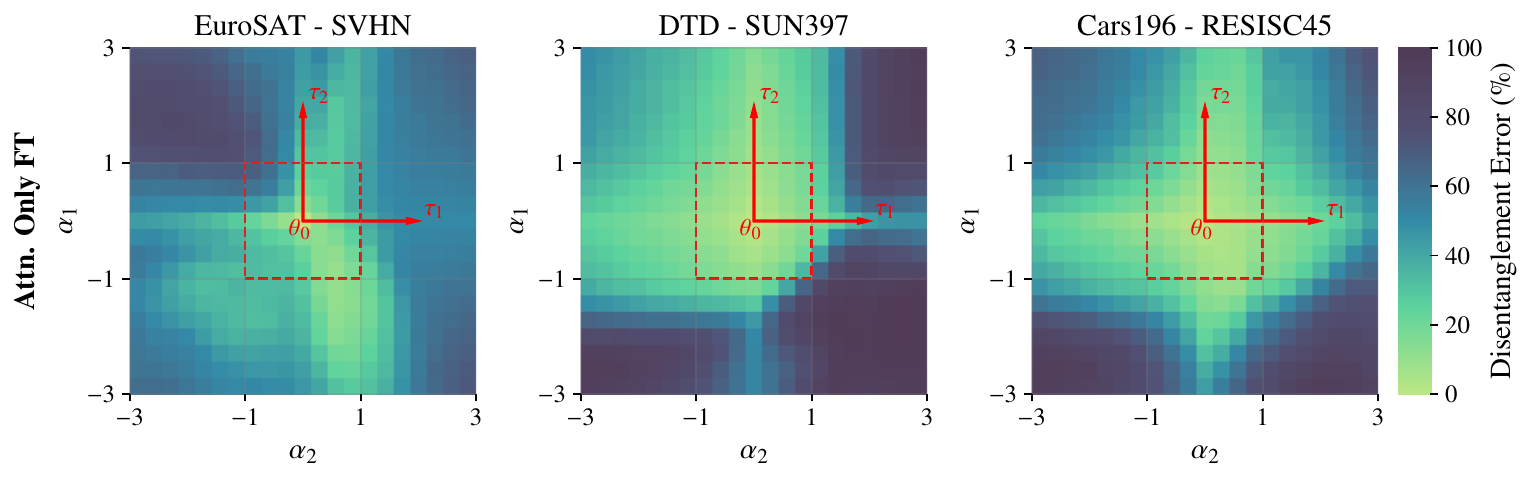}
    \vspace{-1em}
    \includegraphics[width=0.82\linewidth,clip,trim=0 0 0 18]{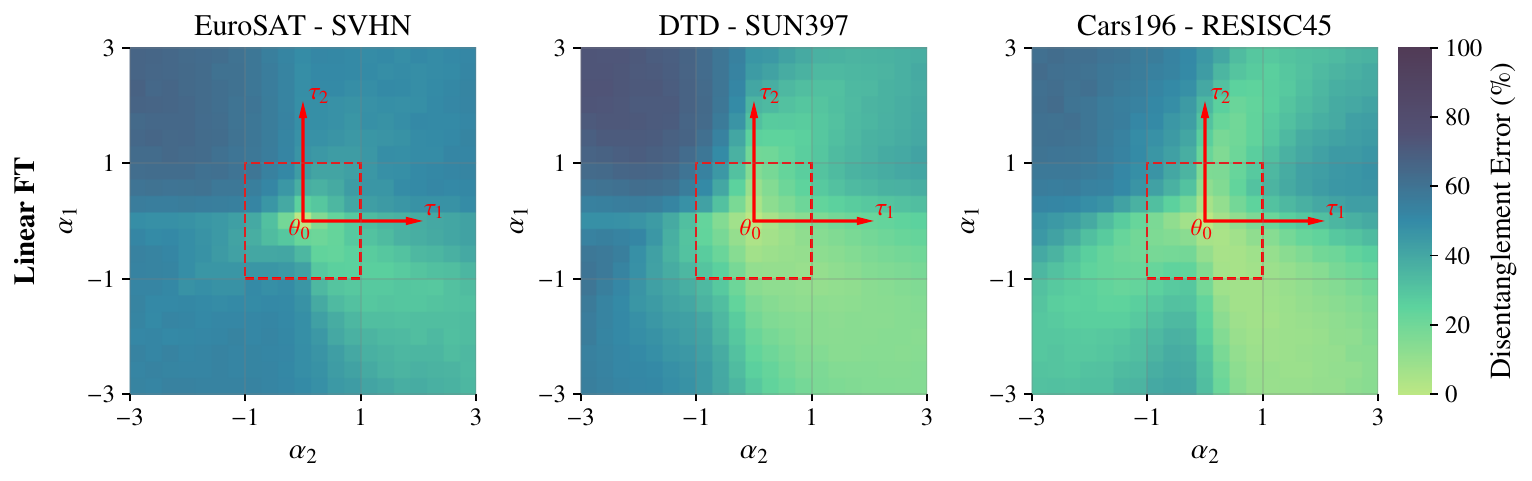}
  \vspace{-1em}
  \includegraphics[width=0.82\linewidth,clip,trim=0 0 0 18]{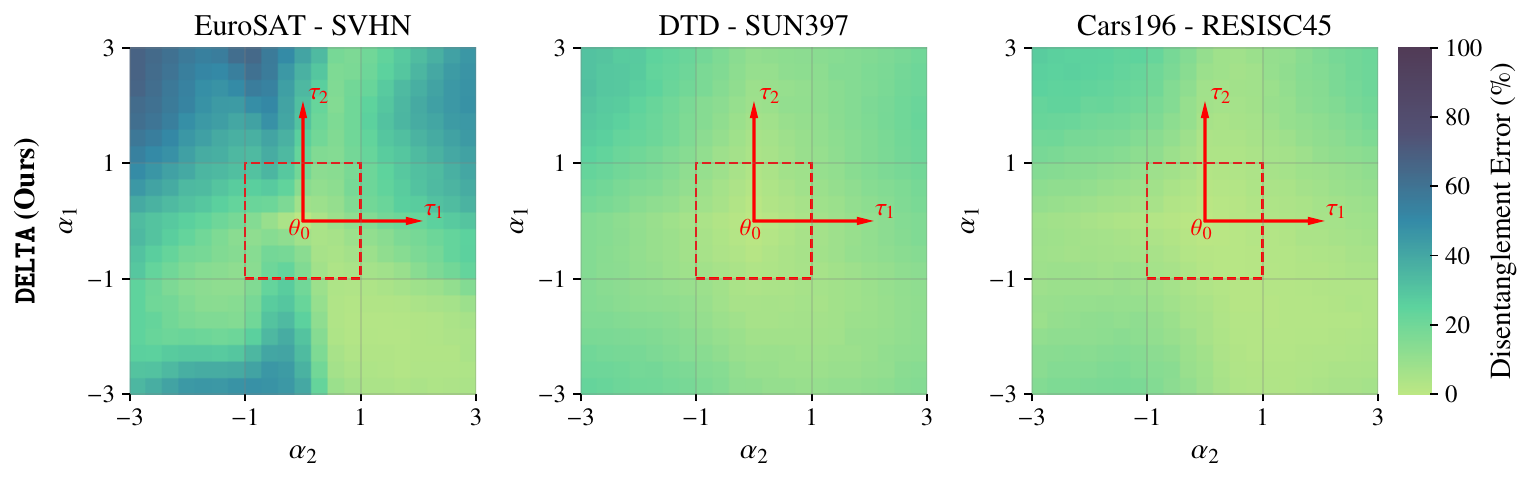}
  \caption{Weight Disentanglement~\cite{ortiz2023task} for Non-Linear FT, Linear FT, Attention-Only FT~\cite{jin2024fine} and our distillation-based approach.}
\label{fig:supp_tareg}
\end{figure}

\nicepar{Sensitivity to the scaling coefficient.} \Cref{fig:lora_sweep_full} reports the merged models' sensitivity to the scaling coefficient, extending our analysis to ViT-L/14 on the \tott{8-Vision} benchmark and ViT-B/32 on the \tott{14-Vision} benchmark. While the peak accuracy of competing methods is occasionally comparable, \methname{} consistently achieves higher overall accuracy across all architectures and benchmarks. Crucially, our method uniquely maintains robust performance across the entire $\alpha$-sweep, effectively eliminating the need for an $\alpha$ grid search.

\begin{figure}
    \centering
    \includegraphics[width=0.95\linewidth]{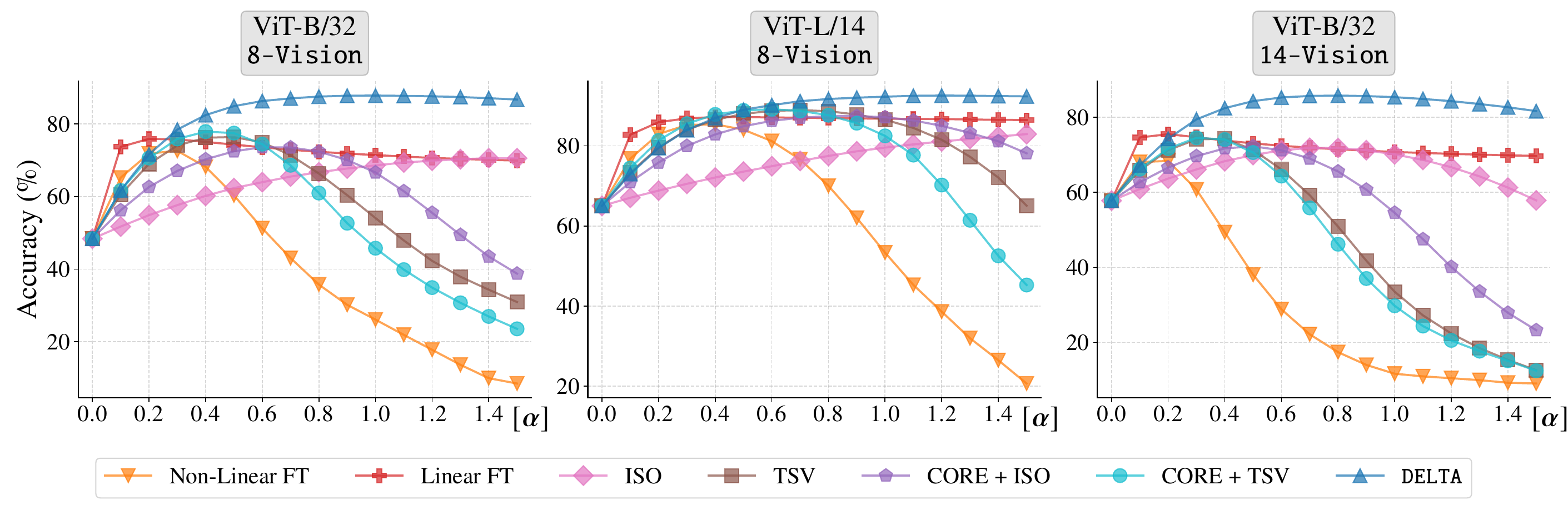}
    \caption{Sensitivity of the merged model to the scaling coefficient on LoRA checkpoints.}
    \label{fig:lora_sweep_full}
\end{figure}

\nicepar{Impact on Task Arithmetic.} In \cref{fig:impact_distillation_full}, we analyze the contribution of each loss component to per-task accuracy, extending our evaluation to ViT-L/14 on the \tott{8-Vision} benchmark and T5-Base on the \tott{6-NLI} benchmark. Additionally, \cref{tab:impact_losses} reports the corresponding aggregate results, covering both task addition and task negation. In the task addition setting, the vision results show that the curvature-regularization objective alone is already highly effective at improving final accuracy. This effect is particularly pronounced for ViT-L/14, where linearization appears to play a more marginal role than in ViT-B/32. A possible explanation is that, due to its larger width, ViT-L/14 may operate closer to an infinite-width training regime, in which the first-order Taylor approximation remains accurate even along naturally non-linear training trajectories. In the textual domain, instead, the results more clearly highlight the complementary role of the two objectives. In the task negation setting, the complementarity between the two objectives becomes even more evident. 

\begin{figure}
    \centering
\includegraphics[width=0.95\linewidth,keepaspectratio]{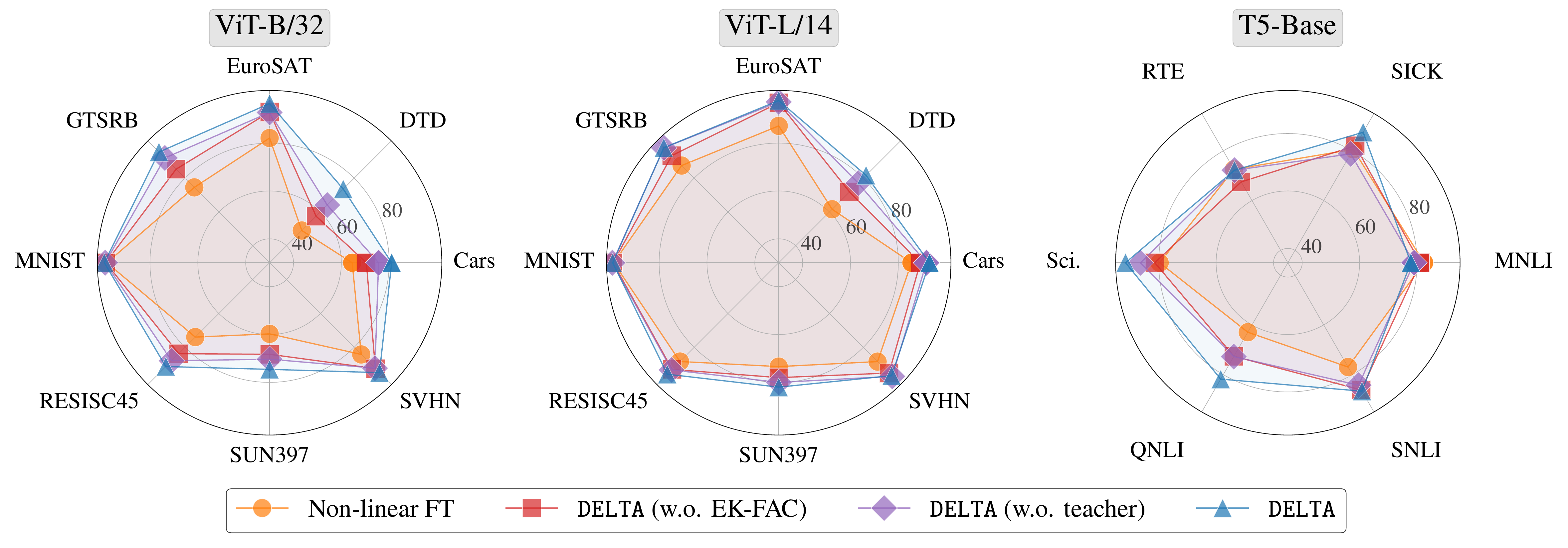}
    \caption{\textbf{Individual task performance.} Accuracy of the merged model on each task, illustrating the impact of our distillation framework across different architectures and benchmarks.}
    \label{fig:impact_distillation_full}
\end{figure}

\begin{table}
\caption{\textbf{Ablation on task addition and task negation.} Left: loss ablation for task addition, reporting the mean absolute accuracy of the merged model across architectures and benchmarks. Right: the same study for task negation, reporting unlearning performance on target tasks and retention on control tasks (ImageNet-1K).}
\label{tab:impact_losses}
\centering

\begin{minipage}{0.34\linewidth}
\centering
\setlength{\tabcolsep}{3pt}
\begin{tabular}{lccc}
\toprule
 & \multicolumn{2}{c}{\textbf{\tott{8-Vision}}} & \textbf{\tott{6-NLI}}\\ 
\cmidrule(lr){2-3} \cmidrule(lr){4-4}
\textbf{Method} & \textbf{B/32} & \textbf{L/14} & \textbf{T5}\\ 
\midrule
Non-Linear FT & $73.5$ & $84.5$ & $78.2$ \\
\midrule
\methname{} (w/o EK-FAC) & $81.1$ & $89.6$ & $78.5$ \\
\methname{} (w/o teacher) & $84.2$ & $91.7$ & $78.9$ \\
\rowcolor{gray!15} \textbf{\methname{}} (ours)
 & $\mathbf{88.3}$ & $\mathbf{92.7}$ & $\mathbf{82.4}$ \\
\bottomrule
\end{tabular}
\end{minipage}
\hfill
\begin{minipage}{0.6\linewidth}
\centering
\setlength{\tabcolsep}{3pt}
\begin{tabular}{lcccccc}
\toprule
& \multicolumn{4}{c}{\textbf{\tott{8-Vision}}} 
& \multicolumn{2}{c}{\textbf{\tott{14-Vision}}}\\
\cmidrule(lr){2-5} \cmidrule(lr){6-7}
\multirow{2}{*}{\textbf{Method}} 
& \multicolumn{2}{c}{\textbf{B/32}}
& \multicolumn{2}{c}{\textbf{L/14}}
& \multicolumn{2}{c}{\textbf{B/32}}\\
& Targ. & Cont. & Targ. & Cont. & Targ. & Cont. \\
\midrule
Non-Linear FT  & $20.4$ & $60.5$ & $18.1$ & $72.3$ & $30.5$ & $60.6$ \\
\midrule
\methname{} (w/o EK-FAC) & $14.2$ & $61.1$ & $13.2$ & $72.9$ & $25.2$ & $61.0$ \\
\methname{} (w/o teacher) & $16.8$ & $62.4$ & $19.0$ & $74.9$ & $25.3$ & $62.2$ \\
\midrule
\rowcolor{gray!15} \rule{0pt}{2ex}
\textbf{\methname{}} (ours)
& $\mathbf{9.6}$ & $\mathbf{62.1}$ & $\mathbf{11.7}$ & $\mathbf{74.7}$ & $\mathbf{19.1}$ & $\mathbf{62.1}$ \\
\bottomrule
\end{tabular}
\end{minipage}

\end{table}

\nicepar{Student-Teacher comparison.} \Cref{fig:suppl_teachervsstudent} reports the single-task performance of the student versus the teacher before merging. In addition to the T5 results on the \tott{6-NLI} benchmark reported in the main paper, we extend this comparison to ViT-L/14 on the \tott{8-Vision} benchmark, as well as ViT-B/32 on both the \tott{8-Vision} and \tott{14-Vision} benchmarks. \methname{} consistently outperforms its linearized teacher across all architectures and datasets. This confirms that the greater expressivity of the non-linear regime is effectively exploited to learn richer representations, rather than simply mimicking the teacher's activations.

\begin{figure}
    \centering
    \includegraphics[width=0.95\linewidth]{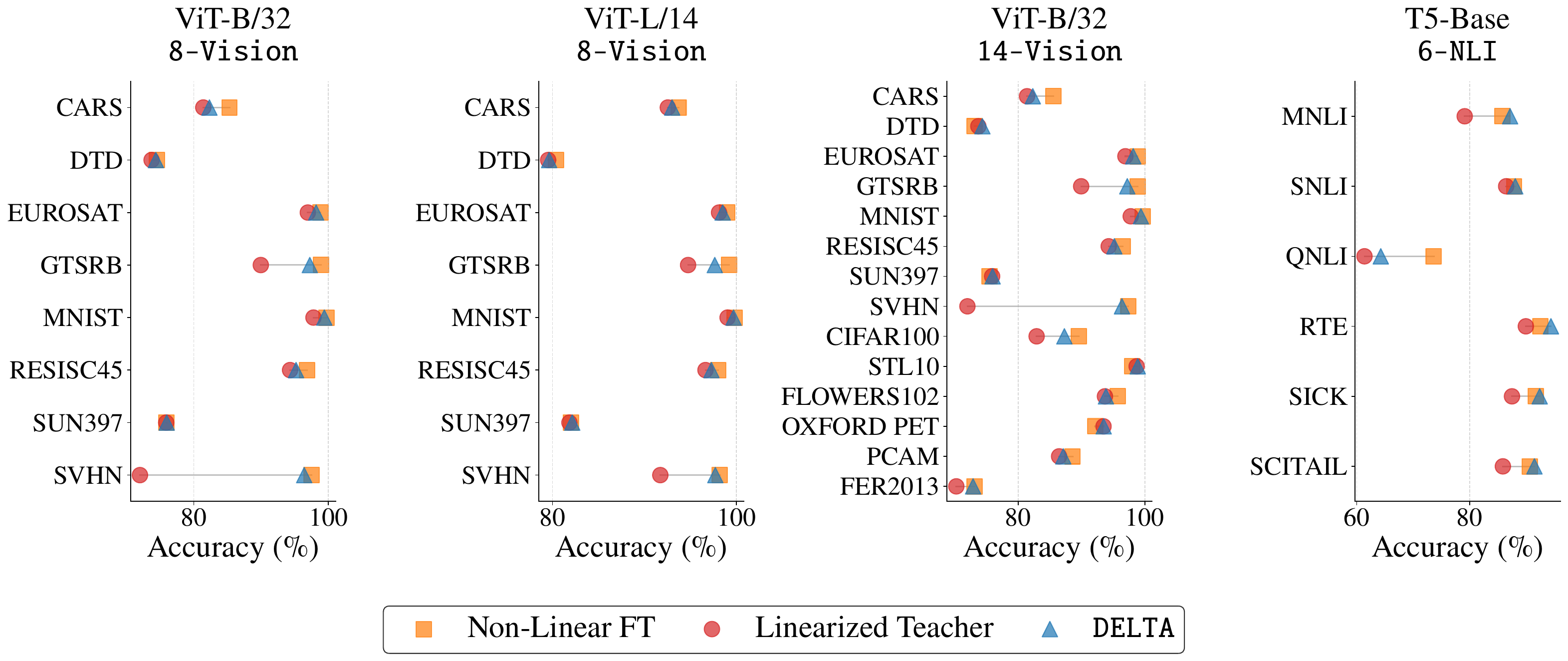}
    
\caption{\changed{\textbf{Student - Teacher comparison. } Accuracy of the single task fine-tuning on the corresponding dataset}}

\label{fig:suppl_teachervsstudent}
\vspace{-0.8em}
\end{figure}

\nicepar{Linearization error.} In \cref{fig:histo_lin_err_14}, we plot the linearization error of our method on the ViT-L/14 backbone for the \tott{8-Vision} benchmark. The results confirm that our full protocol achieves a near-zero linearization error, substantially reducing the discrepancy relative to the non-linear baseline. Our analysis also reveals that teacher distillation is the core driver of this alignment; without it, the linearization error degrades much more severely than when curvature regularization is omitted.

\begin{figure}
    \centering
    \includegraphics[width=\linewidth]{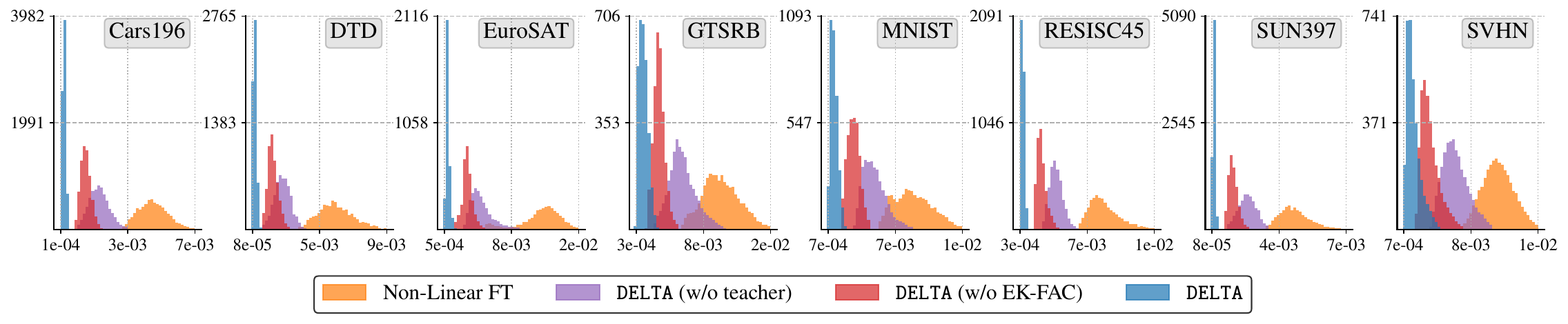}
    \caption{Histogram of the linearization error with the ViT-L/14 backbone.}
    \label{fig:histo_lin_err_14}
\end{figure}

\nicepar{Support localization.} \Cref{fig:histo_mse_suppl} illustrates the impact of each individual loss component on the Mean Squared Error (MSE) between the activations of the pre-trained and individually fine-tuned models. We replicate this analysis for the ViT-L/14 backbone in \cref{fig:histo_mse_suppl_14}. Consistent with the findings in our main text, these ablations confirm that \methname{} achieves \textbf{stronger support localization} than conventional Non-Linear FT. Specifically, the edit distance is substantially higher for In-Domain Data while remaining near zero for Out-of-Domain Data, indicating a strict reversion to the pre-trained model for out-of-distribution inputs.

Crucially, these supplementary results isolate the exact source of this phenomenon: ablating curvature regularization (DELTA w/o Reg) severely disrupts this clean separation, whereas removing distillation (DELTA w/o Distill) largely preserves it. This verifies that curvature regularization, rather than distillation, is the primary driver of support localization.

\begin{figure}[t]
    \centering
    \includegraphics[width=0.92\linewidth]{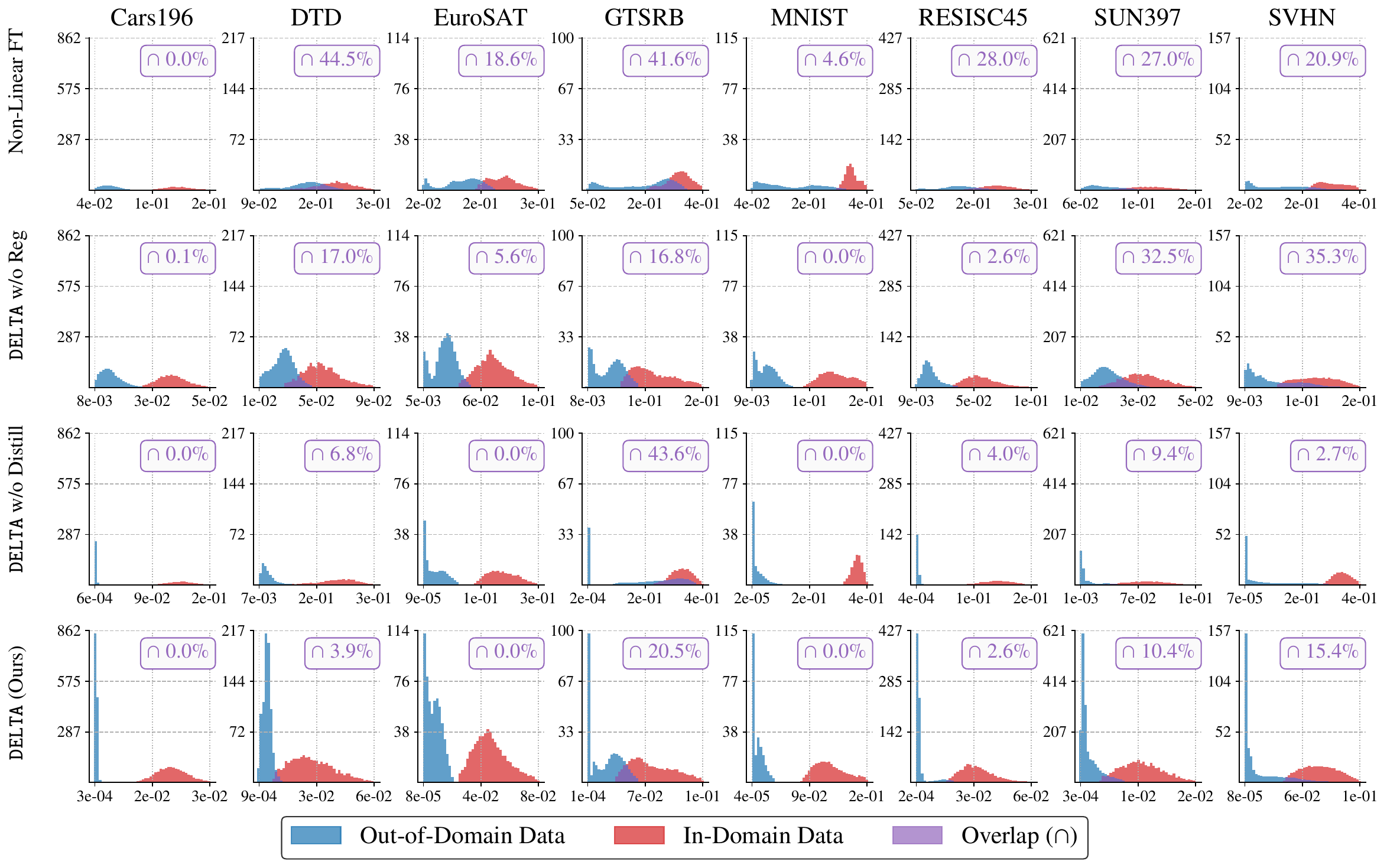}
    \caption{\textbf{Activation MSE Distributions.} Extended histograms of the activation MSE between the pre-trained and fine-tuned models, as defined in \cref{fig:histo_localization}. Evaluated on the ViT-B/32 backbone, we compare our full \methname{} protocol with its ablations (\textit{w/o Distill} and \textit{w/o Reg}) alongside standard Non-Linear FT.}
    \label{fig:histo_mse_suppl}
\end{figure}

\begin{figure*}
    \centering
    \includegraphics[width=0.92\linewidth]{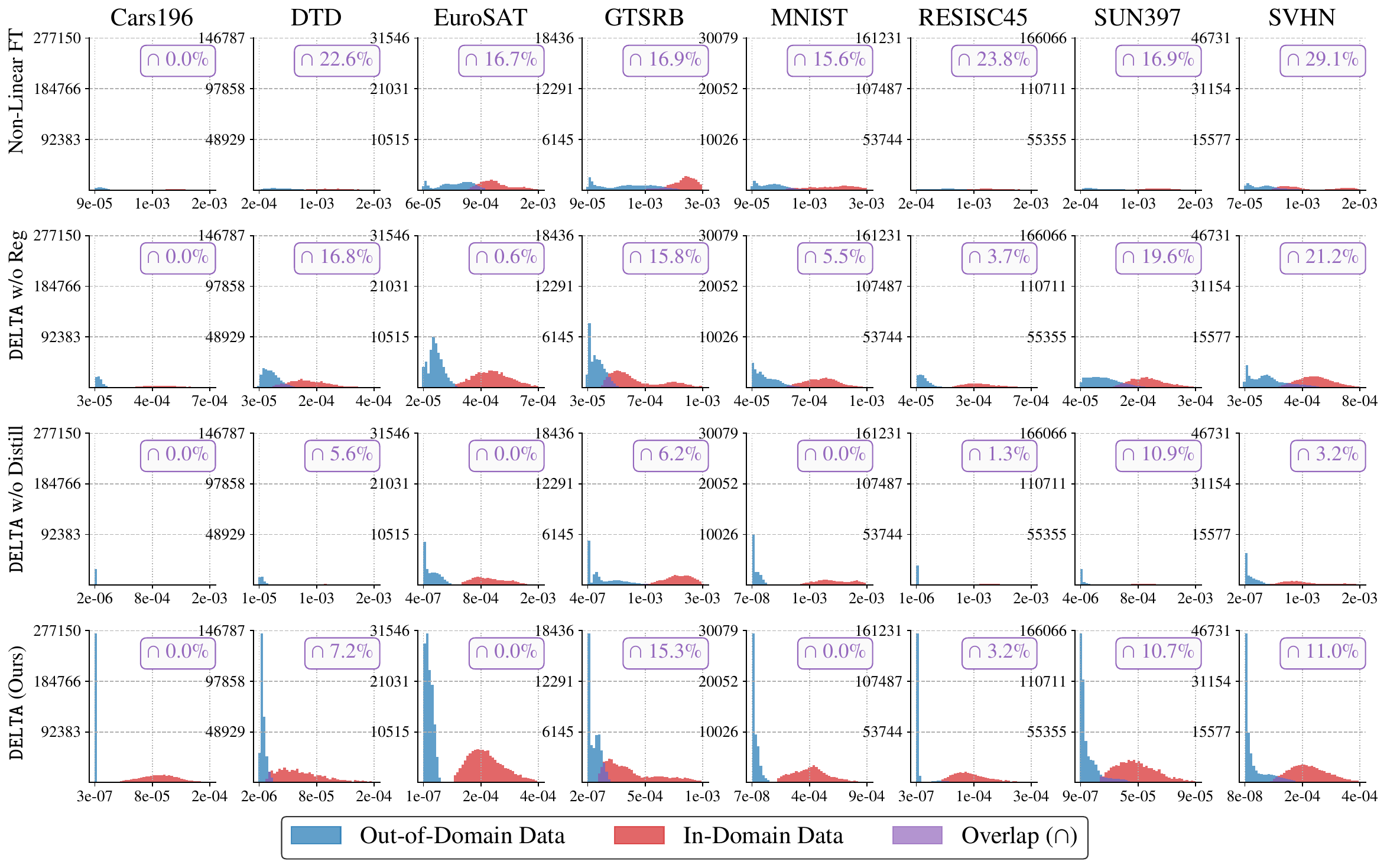}
    \caption{\textbf{Activation MSE Distributions on ViT-L/14.} Histograms of the activation MSE evaluated on the ViT-L/14 backbone. Consistent with the base architecture, the ablation of our loss components confirms that curvature regularization remains the primary driver of support localization at scale.}
    \label{fig:histo_mse_suppl_14}
\vspace{-0.8em}
\end{figure*}

\end{document}